\documentclass[sigconf]{acmart}

\usepackage{balance}
\usepackage{epsfig}
\usepackage{amsmath}
\usepackage{amssymb}
\usepackage[normalem]{ulem}
\usepackage[figuresright]{rotating}
\usepackage{textcomp,subfigure,algorithm,algorithmic,multirow,upgreek,booktabs}
\usepackage{graphics,gensymb,setspace}
\usepackage{caption}
\usepackage{enumitem}
\setitemize{noitemsep,topsep=0pt,parsep=0pt,partopsep=0pt}

\usepackage{wrapfig}
\definecolor{myred}{rgb}{1, 0.2, 0.03}

\usepackage{capt-of}
\usepackage{varwidth}

\def\BibTeX{{\rm B\kern-.05em{\sc i\kern-.025em b}\kern-.08emT\kern-.1667em\lower.7ex\hbox{E}\kern-.125emX}}
    
\copyrightyear{2019} 
\acmYear{2019} 
\acmConference[MM '19]{Proceedings of the 27th ACM International Conference on Multimedia}{October 21--25, 2019}{Nice, France}
\acmBooktitle{Proceedings of the 27th ACM International Conference on Multimedia (MM '19), October 21--25, 2019, Nice, France}
\acmPrice{15.00}
\acmDOI{10.1145/3343031.3350980}
\acmISBN{978-1-4503-6889-6/19/10}

\begin{document}

\fancyhead{}

\title{Cycle In Cycle Generative Adversarial Networks for Keypoint-Guided Image Generation}

\author{Hao Tang}
\email{hao.tang@unitn.it}
\affiliation{%
  \institution{DISI, University of Trento \\ Trento, Italy}
}

\author{Dan Xu}
\email{danxu@robots.ox.ac.uk}
\affiliation{%
  \institution{University of Oxford \\ Oxford, UK}
}

\author{Gaowen Liu}
\email{g_l198@txstate.edu}
\affiliation{%
  \institution{Texas State University 
  \\ San Marcos, USA}
}

\author{Wei Wang}
\email{wei.wang@epfl.ch}
\affiliation{%
  \institution{EPFL \\ Lausanne, Switzerland}
}

\author{Nicu Sebe}
\email{niculae.sebe@unitn.it}
\affiliation{%
  \institution{DISI, University of Trento \\ Trento, Italy}
}

\author{Yan Yan}
\email{y_y34@txstate.edu}
\affiliation{%
  \institution{Texas State University \\ San Marcos, USA}
}


 




%
\renewcommand{\shortauthors}{Trovato and Tobin, et al.}

%

\begin{abstract}
	In this work, we propose a novel Cycle In Cycle Generative Adversarial Network (C$^2$GAN) for the task of keypoint-guided image generation. 
	The proposed C$^2$GAN is a cross-modal framework exploring a joint exploitation of the keypoint and the image data in an interactive manner. 
	C$^2$GAN contains two different types of generators, i.e., keypoint-oriented generator and image-oriented generator. Both of them are mutually connected in an end-to-end learnable fashion and explicitly form three cycled sub-networks, i.e., one image generation cycle and two keypoint generation cycles. Each cycle not only aims at reconstructing the input domain, and also produces useful output involving in the generation of another cycle. By so doing, the cycles constrain each other implicitly, which provides complementary information from the two different modalities and brings extra supervision across cycles, thus facilitating more robust optimization of the whole network. Extensive experimental results on two publicly available datasets, i.e., Radboud Faces~\cite{langner2010presentation} and Market-1501~\cite{zheng2015scalable}, demonstrate that our approach is effective to generate more photo-realistic images compared with state-of-the-art models.
\end{abstract}


%

%

%
\maketitle

\section{Introduction}

Humans have the ability to convert objects or scenes to another form just by imagining, while it is difficult for machines to deal with this task.
For instance, we can easily generate mental images that have different facial expressions and human poses.
In this paper, we study how to enable machines to perform image-to-image translation tasks, which has many application scenarios, such as human-computer interactions, entertainment, virtual reality and data augmentation.
One important benefit of this task is that it can help to augment training data by generating diverse images with given input images, which thus could be employed to improve other recognition or detection tasks. 

\begin{figure}[!t]
	\centering
	\includegraphics[width=1\linewidth]{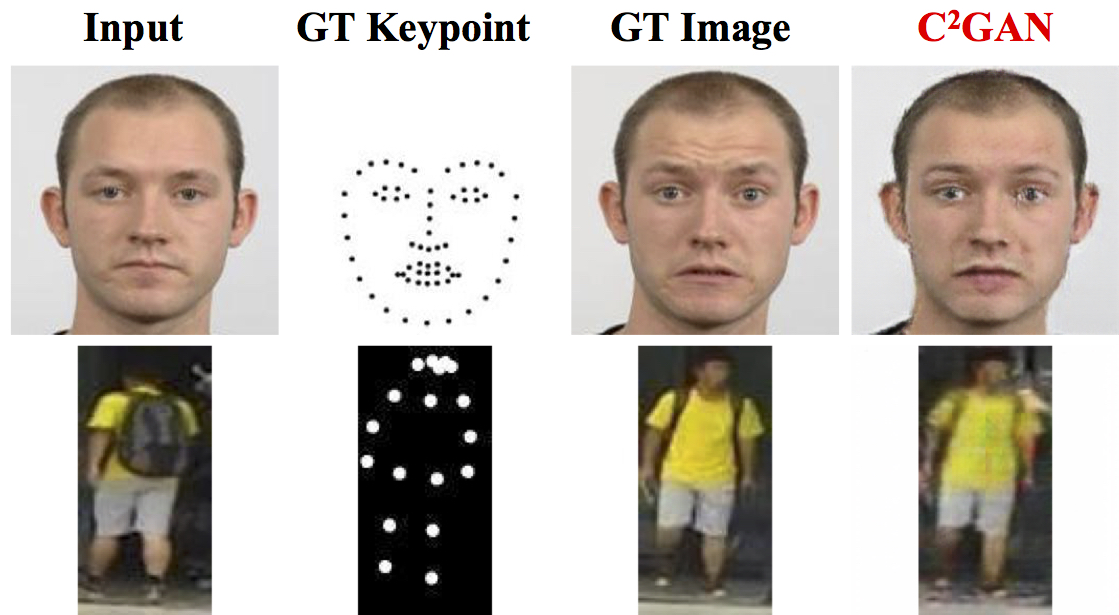}
	\caption{The proposed C$^2$GAN on two challenging tasks, i.e., landmark-guided facial expression generation (Top) and keypoint-guided person pose generation (Bottom). From left to right: input image, ground truth target keypoint, ground truth target image and C$^2$GAN (ours).}
	\label{fig:first_figur}
\end{figure}

However, the task is still challenging since: (i) it needs to handle complex backgrounds with different illumination conditions, objects and occlusions; (ii) it needs a high-level semantic understanding of the mapping between the input images and the output images since the objects in the inputs may have arbitrary poses, sizes, locations and self-occlusions.
Recently, Generative Adversarial Networks (GANs)~\cite{goodfellow2014generative} have shown the potential to solve this difficult task, and it can be utilized, for instance, to convert a face with a neutral expression into different expressions or to transfer a person with a specific pose into different poses.
GANs have produced promising results in many generative tasks, such as photo-realistic image generation~\cite{isola2016image,brock2018large,karras2019style,zhu2017unpaired}, video generation~\cite{chan2018everybody,siarohin2019animating,mathieu2015deep,vondrick2016generating,yan2017skeleton,wang2018video}, text generation~\cite{yu2017seqgan}, audio generation~\cite{oord2016wavenet} and image inpainting~\cite{dolhansky2018eye,zhang2019gazecorrection}.
Recent works have developed powerful image translation systems, e.g., Pix2pix~\cite{isola2016image} and Pix2pixHD~\cite{wang2018pix2pixHD} in supervised settings, where image pairs are required. 
However, paired training data are usually difficult and expensive to obtain. 
To tackle this problem, CycleGAN~\cite{zhu2017unpaired}, DualGAN~\cite{yi2017dualgan} and DiscoGAN~\cite{kim2017learning} provide an interesting insight, in which the models can learn the mapping from one image domain to another with unpaired data.
However, these models encounter the efficiency issue.
For instance, with $m$ different image domains, CycleGAN, DiscoGAN, DualGAN need to train $m(m{-}1)$ generators and discriminators.
While Pix2pix has to train $m(m{-}1)$ generator/discriminator pairs.
Recently, Anoosheh et al. propose ComboGAN~\cite{anoosheh2017combogan}, which only needs to train $m$ generator/discriminator pairs in term of $m$ different image domains.
Tang et al.~\cite{tang2018dual} propose G$^2$GAN, in which a dual-generator and a discriminator can perform unpaired image-to-image translation for multiple domains.
In addition, Choi et al.~\cite{choi2017stargan} propose StarGAN, a single generator/discriminator pair can perform unpaired image-to-image translation for multiple domains.
While the computational complexity of StarGAN is~$\Theta(1)$, this model is not effective in handling some specific image-to-image translation tasks such as person image generation~\cite{ma2017pose,siarohin2017deformable} and hand gesture generation~\cite{tang2018gesturegan}, in which image generation could involve infinity image domains $m$ since human body and hand gesture in the wild can have arbitrary poses, sizes, appearances and locations.

\begin{figure}[!t]
	\centering
	\includegraphics[width=1\linewidth]{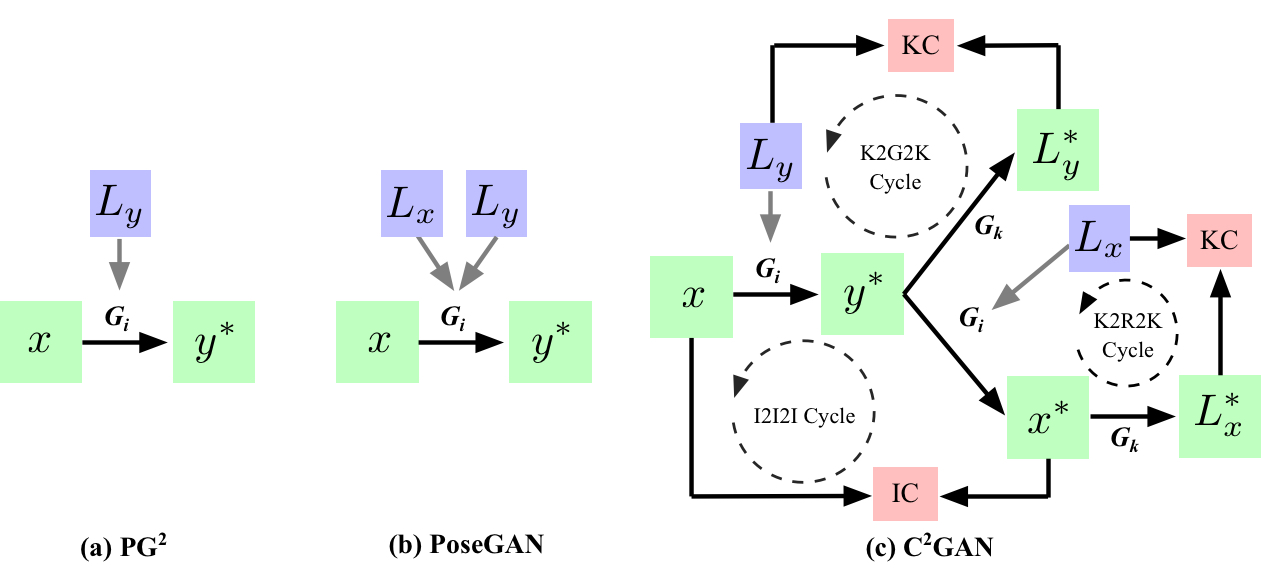}
	\caption{An illustrative comparison of existing models (a) PG$^2$~\cite{ma2017pose}, (b) PoseGAN~\cite{siarohin2017deformable} and (c) the proposed C$^2$GAN. Both PG$^2$ and PoseGAN only use keypoint as input guidance, 
		while C$^2$GAN is a cross-modal model in which keypoint information not only acts as input guidance but also acts as output. The different cycles provide cross-modality complementary information and crossing cycle supervision for better network optimization.
	}
	\label{fig:task}
\end{figure}

To address these limitations, several works are proposed to generate images based on object keypoints or human skeletons. 
Keypoint/skeleton contains the object information of shapes and position, which can be used to produce more photo-realistic images.
For instance, Reed et al.~\cite{reed2016learning} propose GAWWN model, which generates bird images conditioned on both text descriptions and object location.
Qiao et al.~\cite{qiao2018geometry} present GCGAN to generate facial expression conditioned on geometry information of facial landmarks. 
Song et al.~\cite{song2017geometry} propose  G2GAN for facial expression synthesis.
Siarohin et al.~\cite{siarohin2017deformable} introduce PoseGAN for pose-based human image generation.
Tang et al.~\cite{tang2018gesturegan} propose GestureGAN for skeleton-guided hand gesture generation.
Ma et al.~\cite{ma2017pose} propose PG$^2$, which can generate person images using a conditional image and a target pose.
An illustrative comparison among PG$^2$~\cite{ma2017pose}, PoseGAN~\cite{siarohin2017deformable} and the proposed C$^2$GAN is shown in Fig.~\ref{fig:task}.
PG$^2$ tries to generate person images using target keypoints $L_y$. 
For PoseGAN, which needs the target keypoints $L_y$ and original keypoints $L_x$ as conditional inputs. Both methods only employ keypoint information as input guidance. 

Current state-of-the-art keypoint-guided image translation methods such as PG$^2$~\cite{ma2017pose} and PoseGAN~\cite{siarohin2017deformable} have two main issues:
(i) both only directly transfer from an original domain to a target domain, without considering the mutual translation between each other, while the translation across different modalities in a joint network would bring rich cross-modal information.
(ii) both simply employ the keypoint information as input reference to guide the generation, without involving the generated keypoint information as supervisory signals to further improve the network optimization. 
Both issues lead to unsatisfactory results.

To address these limitations, we propose a novel Cycle In Cycle Generative Adversarial Network (C$^2$GAN), in which explicitly three cycled sub-networks are formed to learn the image translation crossing modalities in a unified network structure. 
We have a basic image cycle, i.e., I2I2I ($[x, L_y] \stackrel{G_I}\rightarrow [y^*, L_x] \stackrel{G_I}\rightarrow x^*$), which aims at reconstructing the input and further refine the generated images $y^*$.
The keypoint information in C$^2$GAN is not only utilized as input guidance but also act as output, meaning that the keypoint is also a generative objective. 
Input and output of the keypoint are connected by two keypoint cycles, i.e., K2G2K ($[x, L_y] \stackrel{G_I}\rightarrow y^* \stackrel{G_K}\rightarrow L_y^*$) and K2R2K ($[y^*, L_x] \stackrel{G_I}\rightarrow x^* \stackrel{G_K}\rightarrow L_x^*$), where $G_I$ and $G_K$ denotes an image and a keypoint generator, respectively. 
In this way, keypoint cycles can provide weak supervision to the generated images $y^*$. 
The intuition of the keypoint cycles is that if the generated keypoint is very close to the real keypoint,  then the corresponding images should be similar. In other words, better keypoint generation will boost the image generation, and conversely the improved image generation can facilitate the keypoint generation. These three cycles inherently constraint each other in the network optimization in an end-to-end training fashion.

Moreover, for better optimization the three cycles we propose two novel cycle losses, i.e., Image Cycle-consistency loss (IC) and Keypoint Cycle-consistency loss (KC).  With these cycle losses, each cycle can benefit from each other in joint learning. Moreover, we propose two cross-modal discriminators corresponding to the generators. We conduct extensive experiments on two different keypoint-guided image generation tasks, i.e., landmark-guided facial expression generation and keypoint-guided person pose generation. Extensive experimental results demonstrate that C$^2$GAN yields superior performance compared with state-of-the-art  approaches. 

In summary, the contribution of this paper is three-fold:

\begin{itemize}
	\item We propose a novel cross-modal generative adversarial network named Cycle In Cycle Generative Adversarial Network (C$^2$GAN) for keypoint-guided image generation task, which organizes the keypoint and the image data in an interactive generation manner in a joint deep network, instead of using the keypoint information only as a guidance for the input.
	\item The cycle in cycle structure is a new network design which explores effective utilization of cross-modal information for the keypoint-guided image generation task. The designed cycled sub-networks connect different modalities, and implicitly constraint on each other, leading to extra supervision signals for better image generation. We also investigate cross-modal discriminators and cycle losses for more robust network optimization.
	\item Extensive results on two challenging tasks, i.e., landmark-guided facial expression generation and keypoint-guided person pose generation demonstrate the effectiveness of the proposed C$^2$GAN, and show more photo-realistic image generation compared with existing competing models. 
\end{itemize}

\section{Related Work}

\noindent \textbf{Generative Adversarial Networks (GANs)} \cite{goodfellow2014generative} have shown the capability of generating high-quality images \cite{wang2016generative,karras2017progressive,gulrajani2017improved,brock2018large,karras2019style}.
Although it is successful in many tasks, it also has many challenges, such as how to control the content of the generated images.
To generate meaningful images that meet user requirement, Conditional GAN (CGAN)~\cite{mirza2014conditional} is proposed where the conditioned information is employed to guide the image generation process.
A CGAN model always combines a vanilla GAN and an external information, such as discrete class labels or tags \cite{odena2016semi,perarnau2016invertible,duan2019cascade}, text descriptions \cite{reed2016learning,mansimov2015generating,reed2016generative}
semantic maps \cite{regmi2018cross,wang2018pix2pixHD,tang2019multi,park2019semantic}, conditional images~\cite{isola2016image}, object masks~\cite{mo2018instagan} or attention maps~\cite{tang2019attention,ma2018gan,chen2018attention,mejjati2018unsupervised}.
However, existing CGANs synthesize images based on global constraints such as a class label, text description or facial attribute, they do not provide control over pose, object location or object shape.

\noindent \textbf{Image-to-Image Translation} models use input-output data to learn a parametric mapping between inputs and outputs, e.g., Isola et al.~\cite{isola2016image} propose Pix2pix, which employs a CGAN to learn a mapping function from input to output image domains. 
Wang et al.~\cite{wang2018pix2pixHD} introduce Pix2pixHD model for synthesizing high-resolution images from semantic label maps.
However, most of the tasks in the real world suffer from the constraint of having few or none of the paired input-output samples available.
To overcome this limitation, the unpaired image-to-image translation task has been proposed.
Different from the prior works, unpaired image-to-image translation task learns the mapping function without the requirement of paired training data, such as~\cite{zhu2017unpaired,taigman2016unsupervised,tang2018dual,yi2017dualgan,tang2019attention,kim2017learning,zhou2017genegan,anoosheh2017combogan}.
For instance, Zhu et al.~\cite{zhu2017unpaired} introduce CycleGAN framework, which achieves unpaired image-to-image translation using the cycle-consistency loss.
DualGAN is demonstrated in~\cite{yi2017dualgan}, in which there are image translators to be trained from two unlabeled image sets each representing an image domain.
Kim et al.~\cite{kim2017learning} propose a method based on GANs that learns to discover relations between different domains.

However, existing paired and unpaired image translation approaches are inefficient and ineffective as discussed in the introduction section.
Most importantly, these aforementioned approaches cannot handle some specific image-to-image translation tasks such as person image generation problem~\cite{ma2017pose,siarohin2017deformable}, which could have infinity image domains since a person can have arbitrary poses, sizes, appearances and locations in the wild.

\begin{figure*}[!t]
	\centering
	\includegraphics[width=1\linewidth]{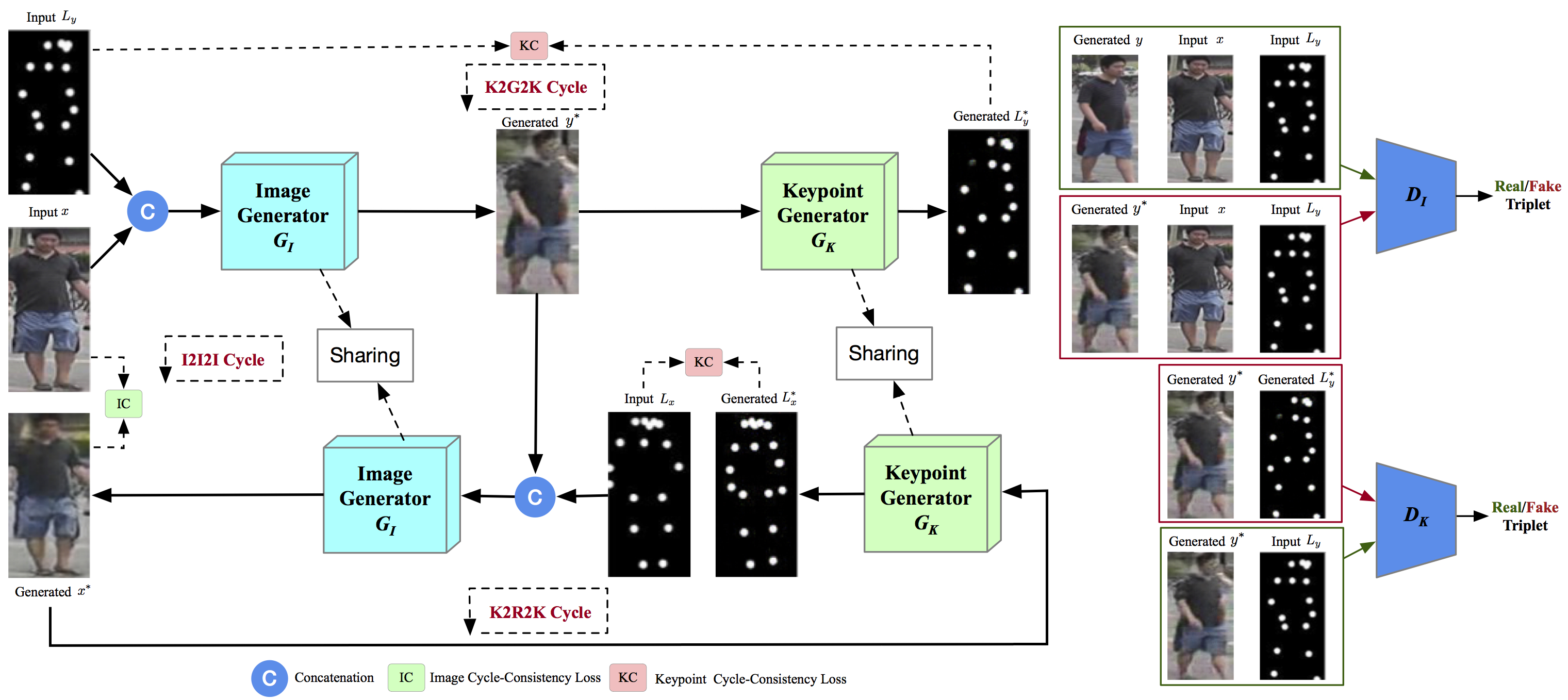}
	\caption{Framework overview of the proposed C$^2$GAN. It contains two types of generators, i.e., image generator $G_I$ and keypoint generator $G_K$.  Parameter-sharing strategies can be used in between the image or the keypoint generators to reduce the model capacity. In the training stage, two generators are explicitly connected by three cycles, i.e., the image cycle I2I2I: $[x, L_y] \stackrel{G_I}\rightarrow [y^*, L_x] \stackrel{G_I}\rightarrow x^*$ and two keypoint cycles K2G2K: $[x, L_y] \stackrel{G_I}\rightarrow y^* \stackrel{G_K}\rightarrow L_y^*$, K2R2K: $[y^*, L_x] \stackrel{G_I}\rightarrow x^* \stackrel{G_K}\rightarrow L_x^*$. The right side of the figure shows different cross-modal discriminators for better network optimization.
	} 
	\label{fig:excyclegan}
\end{figure*}

\noindent \textbf{Keypoint-guided Image-to-Image Translation.}
To address these aforementioned limitations, several works \cite{ma2017pose,siarohin2017deformable,reed2016generating,di2017gp,yan2017skeleton} have been proposed to generate images based on object keypoint.
For instance, Di et al.~\cite{di2017gp} propose GPGAN to synthesize faces based on facial landmarks.
Reed et al.~\cite{reed2016generating} present PixelCNN model to generate images part keypoints and text descriptions.
Korshunova et al.~\cite{korshunova2016fast} use facial keypoints to define the affine transformations of the alignment and realignment steps for face swap. 
Wang et al.~\cite{wang2018every} propose CMM-Net for landmark-guided smile generation. 
Sun et al.~\cite{sun2018natural} propose a two-stage framework to perform
head inpainting conditioned on the generated facial landmark in the first stage.
Chan et al.~\cite{chan2018everybody} propose a method to transfer motion between human subjects based on pose stick figures in different videos.
Yan et al.~\cite{yan2017skeleton} propose a method to generate human motion sequence with a simple background using CGAN and human skeleton information.

The aforementioned approaches focus on a single image generation task.
However, in this paper, we propose a novel Cycle In Cycle Generative Adversarial Network (C$^2$GAN) which is a multi-task model and aims to handle two different tasks using one single network, i.e., image and keypoint generation.
During the training stage, two tasks are restricted mutually by three cycles and benefits from each other.
To the best of our knowledge, the proposed model is the first attempt to generate both the image and the keypoint domain in an interactive generation manner within a unified cycle in cycle GAN framework, for the keypoint-guided image translation task. 
Training GANs are a complicated optimization task and incorporating adversarial keypoint in training provides extra deep supervision to the image generation network compared to using supervision only from the image domain, thus facilitating the network optimization. 
Moreover, the keypoint generation aims not only to approximate the ground truth output but also to fool the discriminator, meaning that the generated keypoints should represent a real face or a person pose. 
The correlations between these keypoints can be learned in the adversarial setting.

\section{Cycle In Cycle GAN (C$^2$GAN)} 

We start to present the proposed Cycle in Cycle Generative Adversarial Network (C$^2$GAN). Firstly, we introduce the network structures of the three different image and keypoint cycles, and also describe details for the corresponding generators and cross-modal discriminators. Secondly, the proposed objective functions for better optimization of the model and also will be illustrated, and finally the implementation details of the whole model and the training procedure are introduced.

\subsection{Model Overview}
The goal of the proposed C$^2$GAN is to learn two different generators in one single network, i.e., keypoint generator and image generator.
Two generators are mutually connected through three generative adversarial cycles, i.e., one image-oriented cycle and two keypoint-oriented cycles.
In the training stage, all the cycled sub-networks are jointly optimized in an end-to-end fashion and each generator benefit from each other due to the richer cross-modal information and the crossing cycle supervision.
The core framework of the proposed C$^2$GAN is illustrated in Fig.~\ref{fig:excyclegan}. In the following, we describe the structure details of the proposed C$^2$GAN.
\subsection{Image-Domain Generative Adversarial Cycle}
\par\textbf{I2I2I Cycle.} The goal of the image cycle I2I2I is to (i) generate image $y^*$ by using the input conditional image $x$ and the target keypoint $L_y$,
and then (ii) reconstruct the input image $x$ by using the generated image $y^*$ and the keypoint $L_x$ of image $x$. 
I2I2I cycle can be formulated as:
\begin{equation}
[x, L_y] \stackrel{G_I}\rightarrow [y^*, L_x] \stackrel{G_I}\rightarrow x^*,
\end{equation}
where $G_I$ is the image generator.
Different from previous works such as PG$^2$~\cite{ma2017pose} and PoseGAN~\cite{siarohin2017deformable}, which only have one mapping $[x, L_y] \stackrel{G_I} \rightarrow y^*$.
StarGAN~\cite{choi2017stargan} uses the target and original domain labels $l_y$ and $l_x$ as condition information to recover the input image.
However, StarGAN can only handle the task which has a specific number of the category. 
For the person pose generation task, which could have infinity image domains since a person in the wild can have arbitrary poses, sizes, appearances and locations. 
In order to solve this limitation, we replace domain labels $l_y$ and $l_x$ in StarGAN by the keypoint $L_y$ and $L_x$.
Follow PG$^2$ we represent the keypoint as heatmaps.
We concatenate $x$ and $L_y$ and feed them into the image generator $G_I$ to generate $y^*$.
Next, we concatenate $y^*$ and $L_x$ as inputs of $G_I$ to reconstruct the original image $x$.
In this way, the forward and backward consistency can be enforcedly further guaranteed.

\noindent \textbf{Image Generator.}
We use the U-net architecture~\cite{ronneberger2015u} for our image generator $G_I$.
U-net is comprised of encoder and decoder with skip connections between them.
We use $G_I$ two times for generating image $y^*$ and reconstructing image $x^*$.
To reduce model capacity, generators $G_I$ shares parameters between image generation and reconstruction.
For image generation, the target of $G_I$ is to generate an image $y^*=G_I(x, L_y)$ conditioned on the target keypoint image $L_y$ which is similar to the real images $y$.
For image reconstruction, the goal of generator $G_I$ is to recover an image $x^*=G_I(y^*, L_x)$ that looks close to the input images $x$.
$G_I$ tries to learn a combined data distribution between the generation and the reconstruction by sharing parameters, which means $G_I$ receives double data in optimization compared to the generators without using parameter sharing strategy.

\noindent \textbf{Cross-modal Image Discriminator.}
Different from previous works such as PG$^2$~\cite{ma2017pose} which employs a single-modal discriminator, we propose a novel cross-modal discriminator which receives both keypoint and image data as input.
$D_I$ receives two images and one keypoint data as input.
More specifically, $D_I$ aims to distinguish between the generated triplet $[x, L_y, G_I(x, L_y)]$ and the real triplet $[x, L_y, y]$ during image generation stage.
We also propose an image adversarial loss $\mathcal{L}^I_{GAN}(G_I, D_I, x, y, L_y)$ based on the vanilla adversarial loss \cite{goodfellow2014generative}.
The image adversarial loss can be formulated as follows:
\begin{equation}
\begin{aligned}
& \mathcal{L}^I_{GAN}(G_I, D_I, x, y, L_y) = \\  
&  \mathbb{E}_{x, L_y, y\sim{p_{\rm data}}(x, L_y, y)}\left[ \log D_I([x, L_y, y])\right]  
+ \\ 
& \mathbb{E}_{x, L_y\sim{p_{\rm data}}(x, L_y)}[\log (1 - D_I([x, L_y, G_I(x, L_y)]))],
\end{aligned}
\label{equ:cgan1}
\end{equation}
$G_I$ tries to minimize $\mathcal{L}^I_{GAN}(G_I, D_I, x, y, L_y)$ while $D_I$ tries to maximize it.
A similar image adversarial loss for image reconstruction mapping $G_I: [y^*, L_x] \rightarrow x^*$ is defined as:
\begin{equation}
\begin{aligned} 
& \mathcal{L}^I_{GAN}(G_I, D_I, x, y, L_x) = \\
&
\mathbb{E}_{x, L_x, y\sim{p_{\rm data}}(x, L_x, y)}[\log D_I([y, L_x, x])]  + \\
& \mathbb{E}_{y^*, L_x, y\sim{p_{\rm data}}(y^*, L_x, y)}[\log (1 - D_I([y, L_x, G_I(y^*, L_x)]))],
\end{aligned}
\label{equ:cgan2}
\end{equation}
where $D_I$ aims at distinguishing between the fake triplet~$[y, L_x, G_I(y^*, L_x)]$ and the real triplet $[y, L_x, x]$.
Thus the overall image adversarial loss is the sum of Eq.~\eqref{equ:cgan1} and Eq.~\eqref{equ:cgan2}:
\begin{equation}
\begin{aligned} 
& \mathcal{L}^I_{GAN}(G_I, D_I, x, y, L_y, L_x) = \\ & \mathcal{L}^I_{GAN}(G_I, D_I, x, y, L_y)  +   \mathcal{L}^I_{GAN}(G_I, D_I, x, y, L_x).
\end{aligned}
\end{equation}

\noindent \textbf{Image Cycle-Consistency Loss.}
To better learn the image cycle I2I2I, we propose an image cycle-consistency loss.
The loss function writes:
\begin{equation}
\begin{aligned}
&  \mathcal{L}^I_{CYC}(G_I, x, L_x, L_y) = \\
&\mathbb{E}_{x, L_x, L_y\sim{p_{\rm data}}(x, L_x, L_y)}[\Arrowvert G_I(G_I(x, L_y), L_x)-x\Arrowvert_1].
\end{aligned}
\label{equ:cycleganloss}
\end{equation}
The reconstructed images $x^*=G_I(G_I(x, L_y), L_x)$ should closely match to the input image $x$.
Note that we use generator $G_I$ two times with the parameter-sharing strategy, and we use $L1$ distance in image cycle-consistency loss to compute a pixel-to-pixel difference between the recovered image $x^*$ and the real input image $x$.

\subsection{Keypoint-Domain Generative Adversarial Cycle}
The motivation of the keypoint cycle is that, if the generated keypoint is similar to the real keypoint then the corresponding two images should be very close, as we can see in Fig.~\ref{fig:excyclegan}.
We have two keypoint cycles K2G2K and K2R2K. Both of them can provide a supervision signal for optimizing better the image cycle.

\noindent\textbf{K2G2K Cycle.}
For the K2G2K cycle, we feed $[x, L_y]$ into the image generator $G_I$ to produce the target image $y^*$.
Then we employ the keypoint generator $G_K$ to produce the keypoint image $L_y^*$ from $y^*$.
The generated keypoint $L_y^*$ should be very close to the real keypoint image $L_y$.
The formulation of K2G2K can be expressed as:
\begin{equation}
\begin{aligned}
& [x, L_y] \stackrel{G_I}\rightarrow y^* \stackrel{G_K}\rightarrow L_y^*.
\end{aligned}
\end{equation}

\noindent\textbf{K2R2K Cycle.}
For K2R2K cycle, the generated image $y^*$ and keypoint image $L_x$ are first concatenated, and then feed into $G_I$ to produce the recovered image $x^*$. We use $G_K$ to generate the keypoint image $L_x^*$ of $x^*$. We assume that the generated keypoint $L_x^*$ is very similar to the real keypoint image $L_x$.
For the K2R2K cycles, it can be formulated as:
\begin{equation}
\begin{aligned}
\quad [y^*, L_x] \stackrel{G_I}\rightarrow x^* \stackrel{G_K}\rightarrow L_x^*.
\end{aligned}
\end{equation}
Both generated keypoints $L_y^*= G_K(G_I(x, L_y))$~and $L_x^* = G_K(G_I(y^*, L_x))$ should have a close match to the input keypoint image $L_y$ and $L_x$, respectively.
Note that the generator $G_K$ could share parameters between the two cycles, i.e., K2G2K and K2R2K.

\noindent\textbf{Keypoint Generator.}
We employ U-net structure~\cite{ronneberger2015u}  for our keypoint generator $G_K$.
The input of $G_K$ is an image and the output is a keypoint representation.
The keypoint generator produces keypoint $L_y^*=G_K(y^*)$ and $L_x^*=G_K(x^*)$ from image $y^*$ and $x^*$, which can provide extra supervision to the image generator. 

\noindent\textbf{Cross-Modal Keypoint Discriminator.}
The proposed keypoint discriminator $D_K$ is a cross-modal discriminator.
It receives both image and keypoint data as inputs.
Thus the keypoint adversarial loss for $D_K$ can be defined as:
\begin{equation}
\begin{aligned}
& \mathcal{L}^K_{GAN}(G_K, D_K, y^*, L_y)  =  \\
&  \mathbb{E}_{y^*, L_y\sim{p_{\rm data}}(y^*, L_y)}\left[ \log D_K([y^*, L_y])\right]  
+ \\ 
& \mathbb{E}_{y^*\sim{p_{\rm data}}(y^*)}[\log (1 - D_K([y^*, G_K(y^*)]))],
\end{aligned}
\label{equ:lcgan1}
\end{equation}
$G_K$ tries to minimize the keypoint adversarial loss $\mathcal{L}^K_{GAN}(G_K, D_K, y^*, L_y)$ while $D_K$ tries to maximize it.
$D_K$ aims to distinguish between the fake pair $[y^*, L_y^*]$ and the real pair $[y^*, L_y]$.
A similar keypoint adversarial loss for the mapping function $G_K: x^* \rightarrow L_x^*$ is defined as:
\begin{equation}
\begin{aligned} 
& \mathcal{L}^K_{GAN}(G_K, D_K, x^*, L_x)  =  \\
&  \mathbb{E}_{x^*, L_x\sim{p_{\rm data}}(x^*, L_x)}\left[ \log D_K([x^*, L_x])\right]  
+ \\ 
& \mathbb{E}_{x^*\sim{p_{\rm data}}(x^*)}[\log (1 - D_K([x^*, G_K(x^*)]))],
\end{aligned}
\label{equ:lcgan2}
\end{equation}
where discriminator $D_K$ aims to distinguish between the fake pair $[x^*, L_x^*]$ and the real pair $[x^*, L_x]$.
Thus, the total keypoint adversarial loss is the sum of Eq.~\eqref{equ:lcgan1} and Eq.~\eqref{equ:lcgan2}:
\begin{equation}
\begin{aligned} 
& \mathcal{L}^K_{GAN}(G_K, D_K, x^*, y^*, L_x, L_y) = \\
& \mathcal{L}^K_{GAN}(G_K, D_K, y^*, L_y) +  \mathcal{L}^K_{GAN}(G_K, D_K, x^*, L_x).
\end{aligned}
\end{equation}

\noindent\textbf{Keypoint Cycle-Consistency Loss.}
To better learn both keypoint cycles, we propose a keypoint cycle-consistency loss, which can be expressed as:
\begin{equation}
\begin{aligned}
& \mathcal{L}^K_{CYC}(G_K, G_I, x, y^*, L_x, L_y) =  \\
& \mathbb{E}_{x, L_y\sim{p_{\rm data}}(x, L_y)}[\Arrowvert G_K(G_I(x, L_y))-L_y\Arrowvert_1] + \\
&  \mathbb{E}_{y^*, L_x\sim{p_{\rm data}}(y^*, L_x)}[\Arrowvert G_K(G_I(y^*, L_x))-L_x\Arrowvert_1].
\end{aligned}
\label{equ:lcycleganloss}
\end{equation}
We use $L1$ distance in the keypoint cycle-consistency loss to compute pixel-to-pixel difference between the generated keypoints $L_x^*$, $L_y^*$ and the real keypoints $L_x$, $L_y$.
During the training stage, the keypoint cycle-consistency loss can backpropagate errors from the keypoint generator to image generator, which facilitates the optimization of the image generator and thus improves the image generation.

\subsection{Joint Optimization Objective}
We also note that pixel loss \cite{siarohin2017deformable,ma2017pose} can be used to reduce changes and constrain generators. 
Thus we adopt the image pixel loss between the real images $y$ and the generated images $y^*$.
We express this loss as:
\begin{equation}
\begin{aligned}
&\mathcal{L}^I_{PIXEL}(G_I, x, L_y, y ) = \mathbb{E}_{x, L_y, y \sim{p_{\rm data}}(x, L_y, y)}[\Arrowvert G_I(x, L_y)-y\Arrowvert_1].
\end{aligned}
\label{equ:pixelloss}
\end{equation}
We adopt $L1$ distance as loss measurement in image pixel loss.
Consequently, the complete objective loss is:
\begin{equation}
\begin{aligned}
& \mathcal{L}(G_I, G_K, D_I, D_K) = \\
& \lambda^I_{gan} * \mathcal{L}^I_{GAN} + \lambda^I_{cyc} * \mathcal{L}^I_{CYC} + \lambda^I_{pixel} * \mathcal{L}^I_{PIXEL} + \\
&
\lambda^K_{gan} * \mathcal{L}^K_{GAN} + \lambda^K_{cyc} * \mathcal{L}^K_{CYC},
\end{aligned}
\label{eqn:allloss}
\end{equation}
where $\lambda^I_{gan}$, $\lambda^I_{cyc}$, $\lambda^I_{pixel}$, $\lambda^K_{gan}$ and $\lambda^K_{cyc}$ are parameters controlling the relative relation of objectives terms.
We aim to solve:
\begin{equation}
\begin{aligned}
G_I^\ast, G_K^\ast
= \arg \mathop{\min}\limits_{\substack{G_I, G_K}} \mathop{\max}\limits_{D_I, D_K} \mathcal{L}(G_I, G_K, D_I, D_K).
\end{aligned}
\end{equation}

\subsection{Implementation Details}
In this section, we introduce the detailed network implementation, the training strategy and the inference.

\noindent\textbf{Network Architecture.} 
For a fair comparison,  we use the U-net architecture in PG$^2$~\cite{ma2017pose} as our generators.
The encoder of generators is built with the basic Convolution-BatchNorm-LReLU layer. 
The decoder of generators is built with the basic Convolution-BatchNorm-ReLU layer. 
The leaky ReLUs in the encoder has a slope 0.2, while all ReLUs in the decoder are not leaky. 
After the last layer, a Tanh function is used.
We employ the PatchGAN discriminator~\cite{isola2016image,zhu2017unpaired} as our discriminators $D_I$ and $D_K$.
The discriminators are built with the basic Convolution-BatchNorm-ReLU layer. 
All ReLUs are leaky, with slope 0.2.
After the last layer, a convolution is applied to map it to a 1-D value, followed by a Sigmoid function.

\noindent\textbf{Training Strategy.} 
We follow the standard optimization method from~\cite{goodfellow2014generative} to optimize the proposed C$^2$GAN, i.e., we alternate between one gradient descent step on $G_I$, $D_I$, $G_K$, and $D_K$, respectively.
The proposed C$^2$GAN is trained end-to-end and can generate image and keypoint image simultaneously, then the generated keypoint will benefit the quality of the generated image.
Moreover, in order to slow down the rate of discriminators $D_I$, $D_K$ relative to generators $G_I$, $G_K$ we divide the objectives by 2 while optimizing discriminators $D_I$, $D_K$.
To enforce discriminators to remember what it has done wrong or right before, we use a history of generated images to update discriminators similar in~\cite{zhu2017unpaired}.
Moreover, we employ OpenFace~\cite{amos2016openface} and OpenPose~\cite{cao2017realtime} to extract keypoint images $L_x$ and $L_y$ on the Radboud Faces and Market-1501 datasets, respectively.
Keypoint of Market-1501 dataset are represented as heatmaps similar as in PG$^2$~\cite{ma2017pose}.
In contrast,  we set the background of the heatmap to white color and the keypoint to black color on Radboud Faces dataset. 

\noindent\textbf{Inference.} 
At inference time, we follow the same settings of PG$^2$~\cite{ma2017pose} and PoseGAN~\cite{siarohin2017deformable} via inputting an image $x$ and a target keypoint $L_y$ into the image generator $G_I$, and then obtain the output target image. 
Similarly, the keypoint generator $G_K$ receives the image $x$ as input and then outputs the keypoint of image~$x$.
We employ the same setting at both training and inference stage.

\section{Experiments}
In this section, we first introduce the details of the datasets used in our experiments, and then we demonstrate the effectiveness of the proposed C$^2$GAN and training strategy by presenting and analyzing qualitative and quantitative results.
\subsection{Experimental Setup}
\label{sec:es}

\textbf{Datasets.} We employ two publicly datasets to validate the proposed C$^2$GAN on two different tasks, including Radboud Faces dataset~\cite{langner2010presentation} for landmark-guided facial expression generation task, and Market-1501 dataset~\cite{zheng2015scalable} for~keypoint-guided person image generation task. 

(i) The Radboud Faces dataset~\cite{langner2010presentation} contains over 8,000 color face images collected from 67 subjects with eight different emotional expressions, i.e., anger, fear, disgust, sadness, happiness, surprise, neutral and contempt.
It contains 1,005 images for each emotion and is captured from five cameras with different angles, and each subject is asked to show three different gaze directions. 
For each emotion, we randomly select 67\% of images as training data and the rest 33\% images as testing data.
Different from StarGAN~\cite{choi2017stargan}, all the images in our experiments are re-scaled to $256 {\times} 256 {\times} 3$ without any pre-processing.
For the landmark-guided facial expression generation task, we need pairs of images of the same face with two different expressions.
We first remove those images in which the face is not detected correctly using the public OpenFace software~\cite{amos2016openface}, leading to 5,628 training image pairs and 1,407 testing image pair.

(ii) The Market-1501 dataset \cite{zheng2015scalable} is a more challenging person re-id dataset and we use it for the person keypoint and person image generation task.
This dataset contains 32,668 images of 1,501 persons captured from six disjoint surveillance cameras. 
Persons vary in pose, illumination, viewpoint and background in this dataset, which makes the person image generation task more challenging.
We follow the setup in PoseGAN~\cite{siarohin2017deformable}. For the training subset, we obtain 263,631 pairs, which is composed of two images of the same person but different poses.
For testing subset, we randomly select 12,000 pairs.
Note that no person overlapping between the training and testing subsets in this dataset.

\begin{figure}[!t]
	\centering
	\includegraphics[width=1\linewidth]{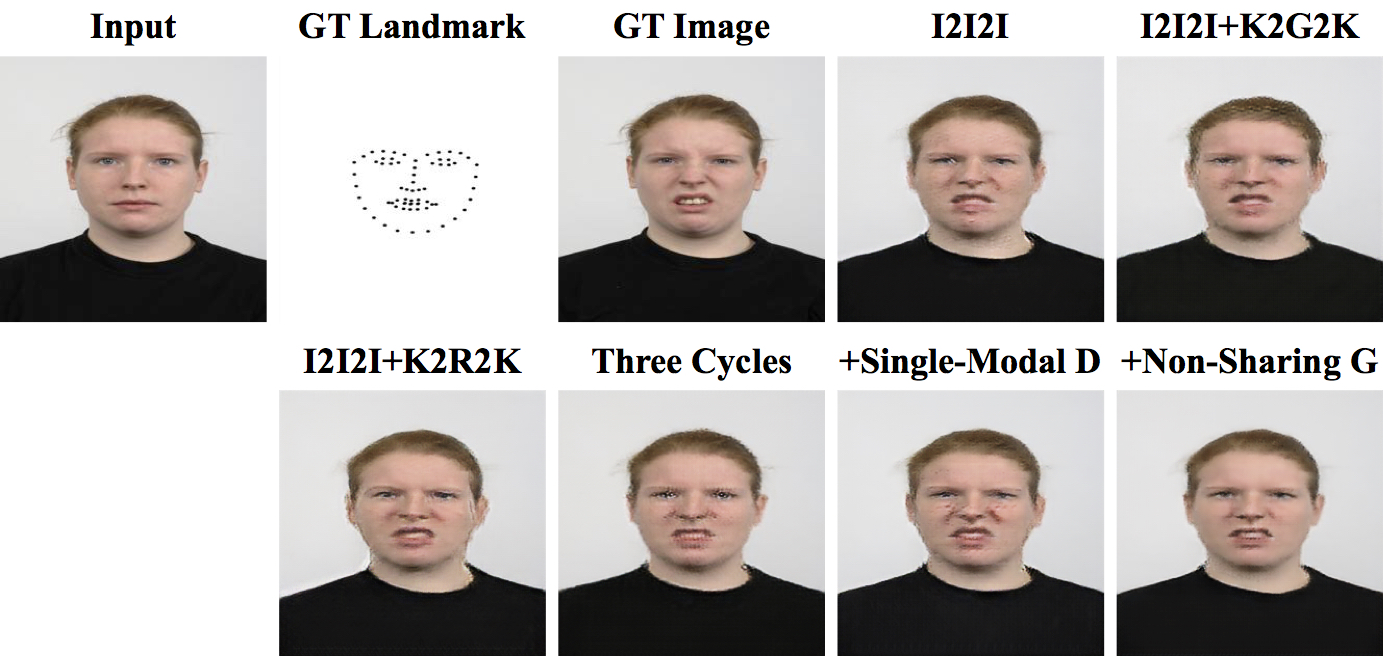}
	\caption{Qualitative results of ablation study on Radbound Faces dataset.}
	\label{fig:abl}
\end{figure}

\begin{table}[!t]
	\centering
	\caption{Quantitative results of ablation study on Radbound Faces dataset. For all metrics, higher is better.} 
	\begin{tabular}{lrrrr}\toprule
		Baseline   &  AMT & PSNR & SSIM   \\ \midrule
		C$^2$GAN w/ I2I2I                      &  25.3 & 21.2030          & 0.8449 \\ 
		C$^2$GAN w/ I2I2I+K2G2K          &   28.2 & 20.8708          & 0.8419 \\ 
		C$^2$GAN w/ I2I2I+K2R2K           & 28.7 & 21.0156          & 0.8437 \\ 
		C$^2$GAN w/ I2I2I+K2G2K+K2R2K & 30.8            & 21.6262          & 0.8540 \\ \hline
		C$^2$GAN w/ Single-Modal D      & 26.4 & 21.2794          & 0.8426 \\ \bottomrule
		C$^2$GAN w/ Non-Sharing G      & \textbf{32.9}& \textbf{21.6353} & \textbf{0.8611} \\ \bottomrule
	\end{tabular}
	\label{tab:abl}
\end{table}

\noindent\textbf{Parameter Setting.}
For both datasets, we do left-right flip for data augmentation similar in PG$^2$~\cite{ma2017pose}.
For optimization, the proposed C$^2$GAN is trained with a batch size of 16 on Radboud Faces dataset.
For a fair comparison, all competing models were trained for 200 epochs on Radboud Faces dataset.
We use the Adam optimizer~\cite{kingma2014adam} with the momentum terms $\beta_1=0.5$ and $\beta_2=0.999$.
The initial learning rate for Adam optimizer is 0.0002. 
For the person image generation task, we train the model for 90 epochs with a smaller batch size 4.

Moreover, we found that the keypoint generator cannot produce accurate keypoints in the early training stage since the image generator produces blurry images during this phase. 
Therefore, we employ a pre-trained OpenPose model~\cite{cao2017realtime}, which replaces the keypoint generator to produce keypoints with location coordinates at the beginning of the training stage.
We also minimize the distance between the generated keypoints (from the generated image) and the corresponding ground truth keypoints (from the ground truth image).
Finally, we incorporate the mask loss proposed in PG$^2$ for person image generation task.

For hyper-parameters setting, we fixed $\lambda^I_{gan}$ and $\lambda^K_{gan}$ to 1 and tune the rest using the grid search.
We found that the weights of reconstruction losses (i.e., $\lambda^I_{cyc}$, $\lambda^I_{pixel}$, $\lambda^K_{cyc}$) set between 10 and 100 yield good performance.
Thus, $\lambda^I_{gan}$, $\lambda^I_{cyc}$, $\lambda^I_{pixel}$, $\lambda^K_{gan}$ and $\lambda^K_{cyc}$ in Eq.~\eqref{eqn:allloss} are set to 1, 10, 10, 1 and 10, respectively.
The proposed C$^2$GAN is implemented using public deep learning framework PyTorch. 
To speed up the training and testing processes, we use an NVIDIA TITAN Xp GPU with 12G memory. 

\noindent\textbf{Evaluation Metric.}
We first adopt AMT perceptual studies to evaluate the quality of the generated images on both datasets similar to~\cite{ma2017pose,siarohin2017deformable}. To seek a quantitative measure that does not require human participation, Structural Similarity (SSIM)~\cite{wang2004image} and Peak Signal-to-Noise Ratio
(PSNR) are employed to evaluate the quantitative quality of generated images on the Radboud Faces dataset.

For Market-1501 dataset on person image generation task, we follow~\cite{siarohin2017deformable} and use SSIM, Inception Score (IS)~\cite{salimans2016improved} and their corresponding masked versions mask-SSIM and mask-IS~\cite{ma2017pose} as measurements.

\subsection{Model Analysis}
We first investigate the effect of the combination of different individual generation cycles to demonstrate the importance of the proposed cycle-in-cycle network structure. Then the parameter-sharing strategy used in the generators for reducing the network capacity is evaluated, and finally the performance influence from the cross-modal discriminators is tested.
All the comparison experiments are conducted via training the models for 50 epochs on Radbound Faces dataset. Fig.~\ref{fig:abl} shows examples of the qualitative results and Table~\ref{tab:abl} shows the quantitative results.

\begin{figure}[!t]
	\centering
	\includegraphics[width=1\linewidth]{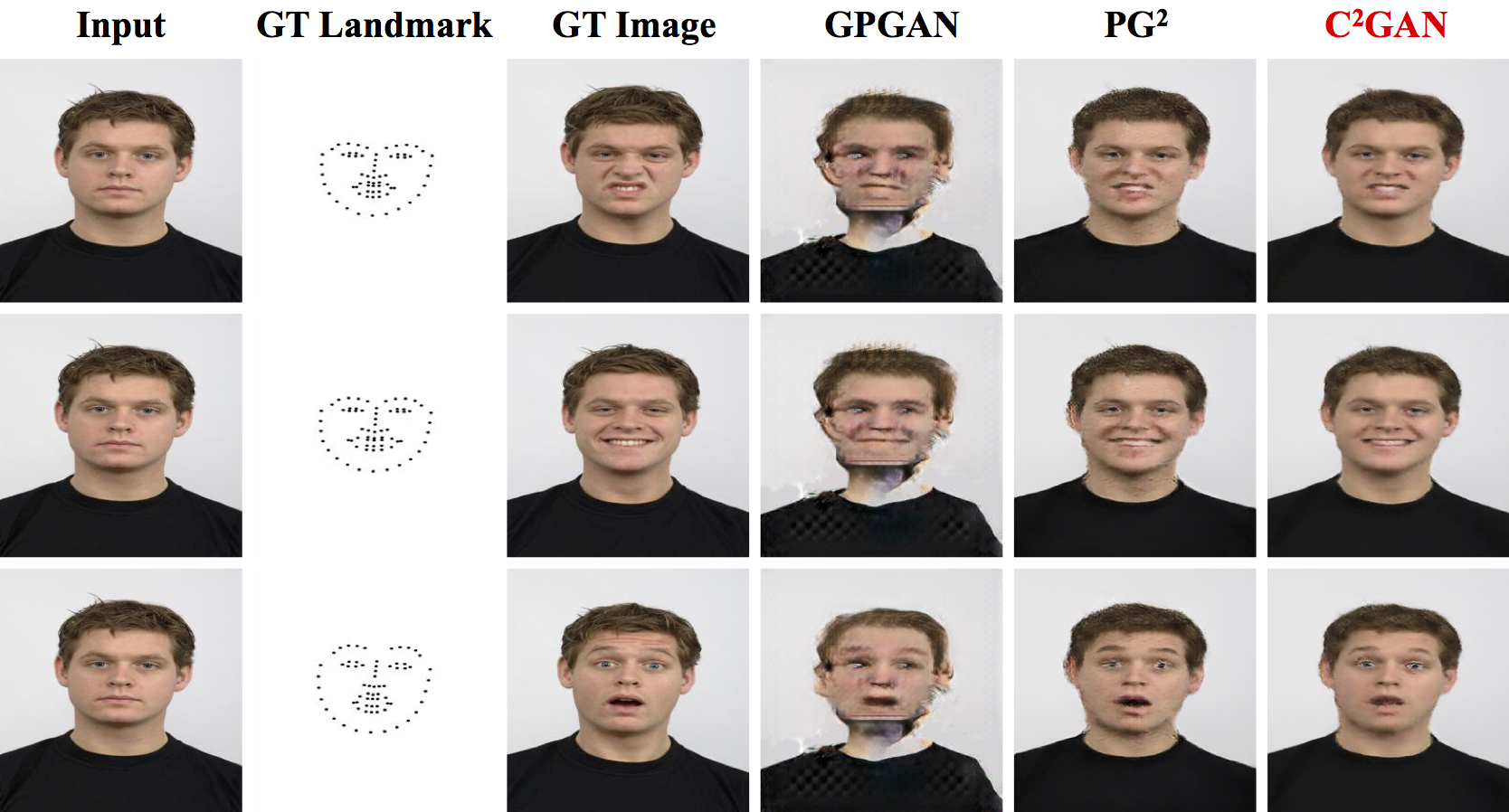}
	\caption{Qualitative comparison with competitors on Radboud Faces dataset. From left to right: input image, ground truth target landmark, ground truth target image, GPGAN~\cite{di2017gp}, PG$^2$~\cite{ma2017pose} and C$^2$GAN (ours).}
	\label{fig:face}
\end{figure}

\begin{table}[!t]
	\centering
	\caption{Quantitative results of different methods on Radboud Faces dataset. For all metrics, higher is better.} 
	\begin{tabular}{lcccccc} \toprule
		Model & Publish & AMT          & SSIM   & PSNR    \\ \midrule		
		GPGAN~\cite{di2017gp} & ICPR 2018 & 0.3          & 0.8185  & 18.7211  \\ 
		PG$^2$~\cite{ma2017pose} & NIPS 2017 &  28.4       & 0.8462 & 20.1462  \\  
		C$^2$GAN (Ours)& - & \textbf{34.2}& \textbf{0.8618}  & \textbf{21.9192} \\  \hline
		Real Data     & - & -            & 1.000  & Inf \\ \bottomrule		
	\end{tabular}
	\label{tab:result_face}
\end{table}	

\begin{figure*}[!t]
	\centering
	\includegraphics[width=0.7\linewidth]{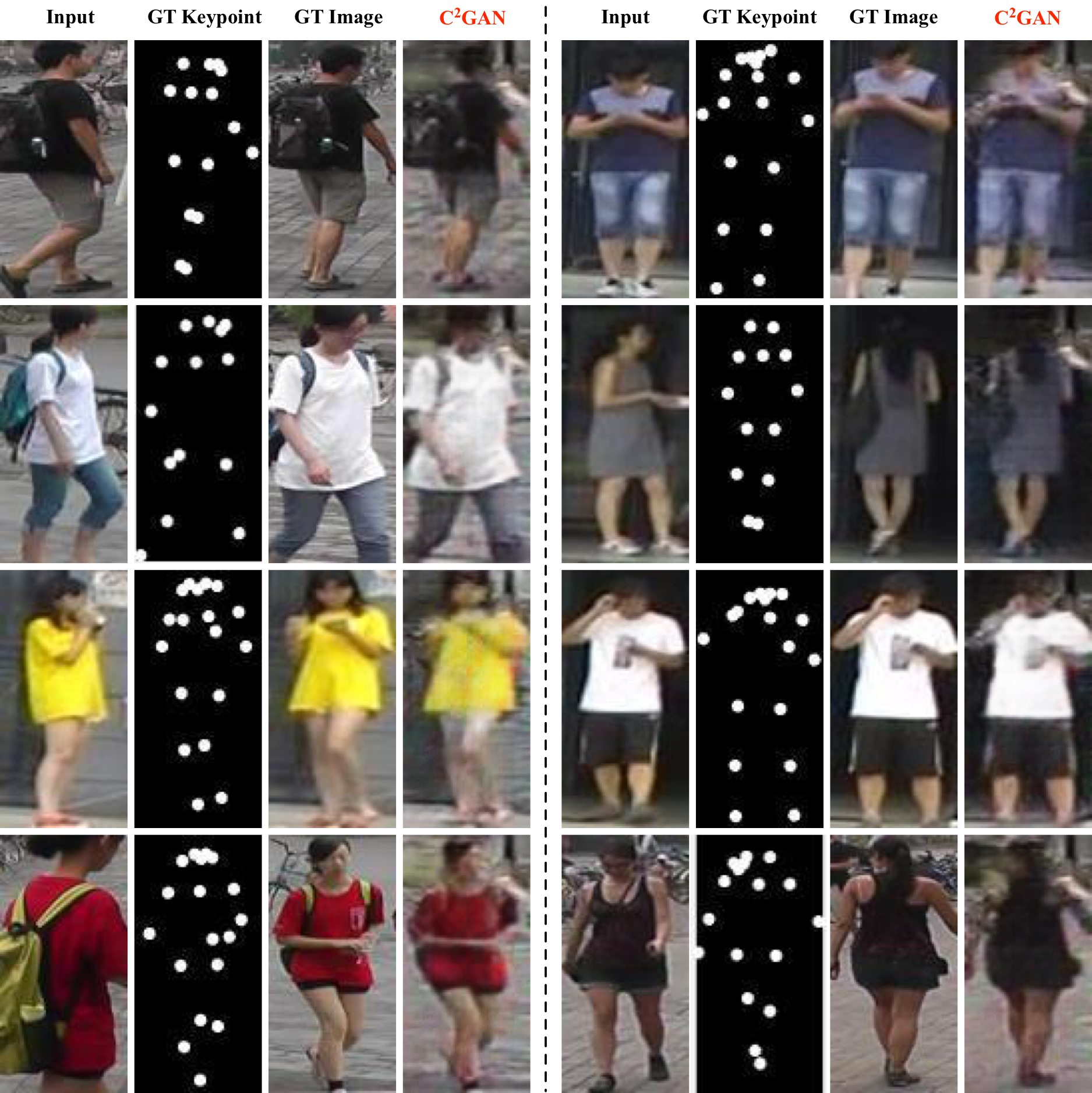}
	\caption{More examples of the qualitative results of the proposed C$^2$GAN on the Market-1501 dataset. From left to right: input image, ground truth target keypoint, ground truth target image and C$^2$GAN (ours).}
	\label{fig:more_pose}
\end{figure*}

\noindent\textbf{Influence of Individual Generation Cycle.}
To evaluate the influence of individual generation cycle, we test with four different combinations of the cycles, i.e., I2I2I, I2I2I+K2G2K, I2I2I+K2R2K, and I2I2I+K2G2K+K2R2K. 
All four baselines use the same training strategies and hyper-parameters. 
As we can see in Table~\ref{tab:abl}, I2I2I, K2G2K and K2R2K are all critical to our final results and the removal of one of them degrades the generation performance, demonstrating our initial intuition that by using cross-modal information in a joint generation framework and by making the cycles constraint on each other boost the final performance. I2I2I+K2G2K+K2R2K obtains the best performance, which is significantly better than the single cycle image network I2I2I, demonstrating the effectiveness of the proposed C$^2$GAN.

\noindent \textbf{Cross-Modal Discriminator vs. Single-Modal Discriminator.}
We also evaluate the performance influence of the proposed cross-modal discriminator (C$^2$GAN w/ I2I2I+K2G2K+K2R2K). Our baseline is the traditional single-modal discriminator (C$^2$GAN w/ Single-Modal D). 
The single modal-D means the discriminator receives only images as input, i.e., the real input images and the generated images. 
From Table~\ref{tab:abl}, it is clear that the proposed cross-modal discriminator performs better than the single-modal discriminator on all evaluation metrics, meaning that the rich cross-modal information could help to learn better discriminator and thus facilitate the optimization of the generator.

\noindent \textbf{Parameter Sharing between Generators.}
The parameter sharing could remarkably reduce the parameters of the whole network. 
We further evaluate how parameter sharing would affect generation performance. 
We test two different baselines: one is C$^2$GAN w/ I2I2I+K2G2K+K2R2K, which shares the parameters of the two image generators $G_I$ and the two keypoint generators $G_K$, respectively. While  C$^2$GAN w/ Non-Sharing G separately learns the four generators. We can observe from Table~\ref{tab:abl} that the non-sharing one achieves slightly better performance than sharing one. However, the number of parameters of non-sharing one is 217.6M, which doubles that of the sharing one. It means that the parameter sharing is a good strategy for balancing performance and overhead. 

\begin{figure*}[!t]
	\centering
	\includegraphics[width=0.84\linewidth]{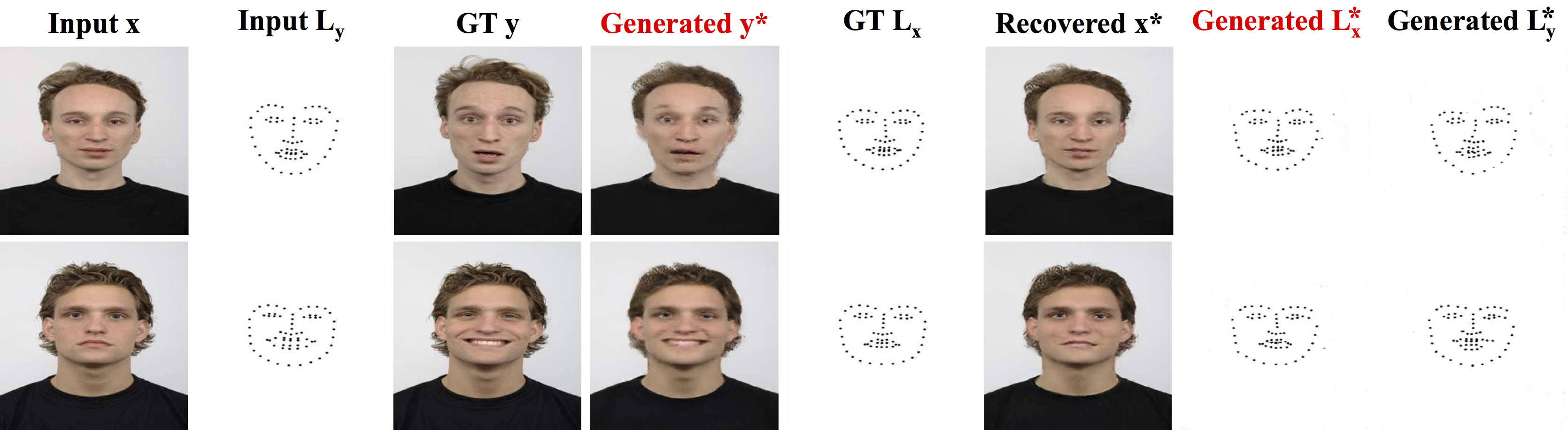}
	\caption{Visualization of keypoint generation on the Radboud Faces dataset.}
	\label{fig:keypoint_visualization}
\end{figure*}

\begin{table*}[!t]
	\centering
	\caption{Quantitative results of different models on Market-1501 dataset. For all the metrics, higher is better.}
	\begin{tabular}{lccccccc} \toprule
		Model & Publish &  AMT (R2G)     & AMT (G2R)     & SSIM           & IS             & mask-SSIM      & mask-IS \\ \midrule 	
		PG$^2$~\cite{ma2017pose} & NIPS 2017                  &  11.2          & 5.5           & 0.253          & 3.460          & 0.792          & 3.435   \\ 
		DPIG~\cite{ma2018disentangled} & CVPR 2018               &  -             & -             & 0.099          & \textbf{3.483} & 0.614          & 3.491   \\  
		PoseGAN~\cite{siarohin2017deformable} & CVPR 2018  &  22.7        & \textbf{50.2}  &\textbf{0.290} & 3.185          & 0.805          & 3.502   \\ 
		C$^2$GAN (Ours) & -                    &  \textbf{23.2} & 46.7          & 0.282          & 3.349          & \textbf{0.811} & \textbf{3.510}   \\ \hline
		Real Data                           & -& -              & -             & 1.000          & 3.860          & 1.000          & 3.360  \\ \bottomrule 
	\end{tabular}
	\label{tab:result_market}
\end{table*} 

\begin{figure}[!t]
	\centering
	\includegraphics[width=1\linewidth]{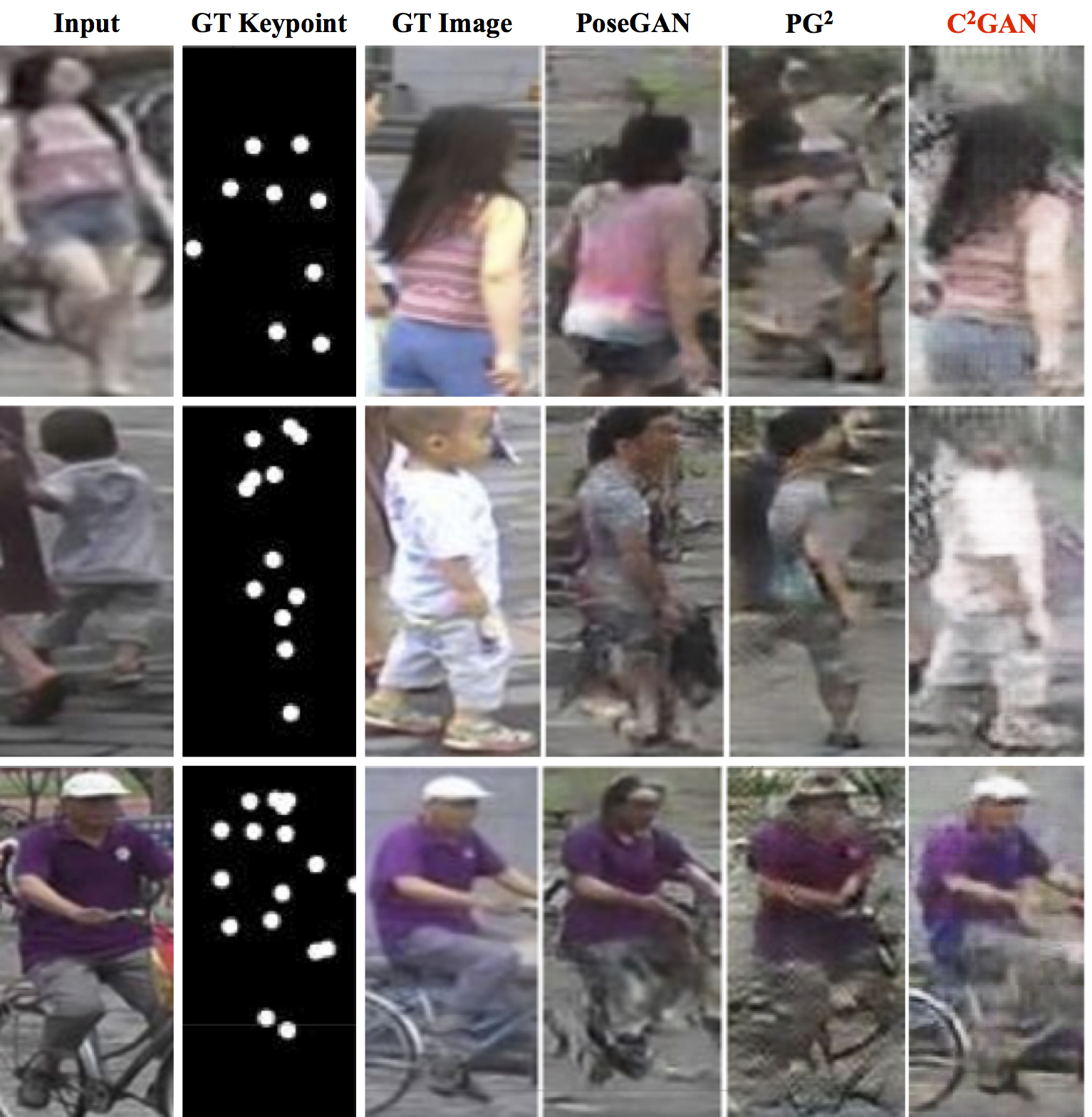}
	\caption{Qualitative comparison with PG$^2$ and PoseGAN on the Market-1501 dataset. From left to right: input image, ground truth target keypoint, ground truth target image, PoseGAN~\cite{siarohin2017deformable}, PG$^2$~\cite{ma2017pose} and C$^2$GAN (ours).}
	\label{fig:result_pose}
\end{figure}

\subsection{Comparison against the State-of-the-Art}

\noindent \textbf{Competing Models.} We consider several state-of-the-art keypoint-guided image generation models as our competitors, i.e.,
GPGAN~\cite{di2017gp}, PG$^2$~\cite{ma2017pose}, DPIG~\cite{ma2018disentangled} and PoseGAN~\cite{siarohin2017deformable}.
Note that for PG$^2$, DPIG and PoseGAN, all of them need to use off-the-shelf keypoint detection models to extract the keypoints, and all of them are restricted on dataset sources or tasks where the keypoint information is available. 
Different from PG$^2$, DPIG and PoseGAN, which focus on the person image generation task, the proposed C$^2$GAN is a general model and learns image and keypoint generation simultaneously in a joint network.
For a fair comparison, we implement all the models using the same setups as our approach.

\noindent \textbf{Task 1: Landmark-Guided Facial Expression Generation.} 
A qualitative comparison of different models on the Radboud Faces dataset is shown in Fig.~\ref{fig:face}.
It is clear that GPGAN performs the worse among all the comparison models.
While the results of PG$^2$ tend to be blurry.
Compared with both GPGAN and PG$^2$, the results of C$^2$GAN are more smooth, sharper and contains more image details. 

Existing competitors such as PG$^2$~\cite{ma2017pose} conduct the experiments on
256$\times$256 resolution images. 
For a fair comparison with them, we also conduct experiments on this image resolution size but note that the proposed C$^2$GAN can be applied for any size images with only small architecture modification.

\noindent \textbf{Task 2: Keypoint-Guided Person Image Generation.}
Fig.~\ref{fig:result_pose} shows the results of PG$^2$, PoseGAN and C$^2$GAN on the Market-1501 dataset.
As we can see, the proposed C$^2$GAN is able to generate visually better images than PG$^2$ and PoseGAN. 
For the first row of results in Fig.~\ref{fig:result_pose}, C$^2$GAN can generate reasonable results while PG$^2$ cannot produce any meaningful content. 
PoseGAN can generate the person but cannot preserve the color information.
For the second row of results in Fig.~\ref{fig:result_pose}, both PoseGAN and PG$^2$ failed to generate the same child, while C$^2$GAN can generate the same child with only a small part missing at the head.
For the last row of results in Fig.~\ref{fig:result_pose}, we can clearly observe the advantage of C$^2$GAN as both PoseGAN and PG$^2$ failed to generate the hat.
We also provide more qualitative results of C$^2$GAN in Fig.~\ref{fig:more_pose}.
As we can see that the proposed C$^2$GAN can generate photo-realistic images with convincing  details. 
Moreover, the generated images are very close to the ground truths. 

Finally, we also note that GANs are difficult to train and easily have mode collapse. However, in keypoint-guided image generation tasks, avoiding mode collapse is not necessarily needed since if you input a person image and a target pose, the model tries to generate this particular person in this particular pose.

\noindent \textbf{Quantitative Comparison of Both Tasks.}
We provide here quantitative results and analysis on both tasks. 
As shown in Table~\ref{tab:result_face}, C$^2$GAN achieves the best performance on the Radboud Faces dataset with all the metrics for landmark-guided facial expression generation task.
Moreover, we quantitatively compare the proposed C$^2$GAN with PoseGAN, DPIG and PG$^2$  on keypoint-guided person image generation task in Table~\ref{tab:result_market}. 
We can observe that C$^2$GAN obtains better performance than PG$^2$ and DPIG on all the evaluation metrics except for IS. Compared with PoseGAN, C$^2$GAN yields very competitive performance.
Specifically, we achieve better performance in terms of the AMT (R2G), IS, mask-SSIM and mask-IS metrics.  

\subsection{Visualization of Keypoint Generation}
C$^2$GAN is a cross-modal generation model and it is not only able to produce the target person but also able to produce the keypoint of the input image.
Both generation tasks benefit from the improvement of each other in an end-to-end training fashion.
We present examples of the keypoint generation results on Radboud Faces dataset in Fig.~\ref{fig:keypoint_visualization}. The inputs are image $x$ and keypoint $L_y$ and the outputs are image $y^*$ and keypoint $L_x^*$, other images and keypoints are given for comparison. As we can see that the generated keypoint $L_x^*$ is very close to the real keypoint $L_x$, which verifies the effectiveness of the keypoint generator~$G_k$ and the joint learning strategy.

\section{Conclusion}
In this paper, we propose a novel Cycle In Cycle Generative Adversarial Network (C$^2$GAN) for keypoint-guide image generation task. 
C$^2$GAN contains two different types of generators, i.e., keypoint-oriented generator and image-oriented generator. The image generator aims at reconstructing the target image based on a conditional image and the target keypoint, and the keypoint generator tries to generate the target keypoint and further provide cycle supervision to the image generator for generating more photo-realistic images. Both generators are connected in a unified network and can be optimized in an end-to-end fashion. Both qualitative and quantitative experimental results on facial expression and person pose generation tasks demonstrate that our proposed framework is effective to generate high-quality images with convincing details.

\section*{Acknowledgments}
We want to thank the NVIDIA Corporation for the donation of the
TITAN Xp GPUs used in this work.

\vfill\eject
\bibliographystyle{ACM-Reference-Format}
\bibliography{sample-bibliography}


\begin{thebibliography}{00}


\ifx \showCODEN    \undefined \def \showCODEN     #1{\unskip}     \fi
\ifx \showDOI      \undefined \def \showDOI       #1{#1}\fi
\ifx \showISBNx    \undefined \def \showISBNx     #1{\unskip}     \fi
\ifx \showISBNxiii \undefined \def \showISBNxiii  #1{\unskip}     \fi
\ifx \showISSN     \undefined \def \showISSN      #1{\unskip}     \fi
\ifx \showLCCN     \undefined \def \showLCCN      #1{\unskip}     \fi
\ifx \shownote     \undefined \def \shownote      #1{#1}          \fi
\ifx \showarticletitle \undefined \def \showarticletitle #1{#1}   \fi
\ifx \showURL      \undefined \def \showURL       {\relax}        \fi
\providecommand\bibfield[2]{#2}
\providecommand\bibinfo[2]{#2}
\providecommand\natexlab[1]{#1}
\providecommand\showeprint[2][]{arXiv:#2}

\bibitem[\protect\citeauthoryear{Amos, Ludwiczuk, and Satyanarayanan}{Amos
  et~al\mbox{.}}{2016}]%
        {amos2016openface}
\bibfield{author}{\bibinfo{person}{Brandon Amos}, \bibinfo{person}{Bartosz
  Ludwiczuk}, {and} \bibinfo{person}{Mahadev Satyanarayanan}.}
  \bibinfo{year}{2016}\natexlab{}.
\newblock \bibinfo{booktitle}{{\em OpenFace: A general-purpose face recognition
  library with mobile applications}}.
\newblock \bibinfo{type}{{T}echnical {R}eport}.
  \bibinfo{institution}{CMU-CS-16-118, CMU School of Computer Science}.
\newblock


\bibitem[\protect\citeauthoryear{Anoosheh, Agustsson, Timofte, and
  Van~Gool}{Anoosheh et~al\mbox{.}}{2018}]%
        {anoosheh2017combogan}
\bibfield{author}{\bibinfo{person}{Asha Anoosheh}, \bibinfo{person}{Eirikur
  Agustsson}, \bibinfo{person}{Radu Timofte}, {and} \bibinfo{person}{Luc
  Van~Gool}.} \bibinfo{year}{2018}\natexlab{}.
\newblock \showarticletitle{ComboGAN: Unrestrained Scalability for Image Domain
  Translation}. In \bibinfo{booktitle}{{\em CVPR Workshop}}.
\newblock


\bibitem[\protect\citeauthoryear{Brock, Donahue, and Simonyan}{Brock
  et~al\mbox{.}}{2018}]%
        {brock2018large}
\bibfield{author}{\bibinfo{person}{Andrew Brock}, \bibinfo{person}{Jeff
  Donahue}, {and} \bibinfo{person}{Karen Simonyan}.}
  \bibinfo{year}{2018}\natexlab{}.
\newblock \showarticletitle{Large scale gan training for high fidelity natural
  image synthesis}. In \bibinfo{booktitle}{{\em ICLR}}.
\newblock


\bibitem[\protect\citeauthoryear{Cao, Simon, Wei, and Sheikh}{Cao
  et~al\mbox{.}}{2017}]%
        {cao2017realtime}
\bibfield{author}{\bibinfo{person}{Zhe Cao}, \bibinfo{person}{Tomas Simon},
  \bibinfo{person}{Shih-En Wei}, {and} \bibinfo{person}{Yaser Sheikh}.}
  \bibinfo{year}{2017}\natexlab{}.
\newblock \showarticletitle{Realtime multi-person 2d pose estimation using part
  affinity fields}. In \bibinfo{booktitle}{{\em CVPR}}.
\newblock


\bibitem[\protect\citeauthoryear{Chan, Ginosar, Zhou, and Efros}{Chan
  et~al\mbox{.}}{2018}]%
        {chan2018everybody}
\bibfield{author}{\bibinfo{person}{Caroline Chan}, \bibinfo{person}{Shiry
  Ginosar}, \bibinfo{person}{Tinghui Zhou}, {and} \bibinfo{person}{Alexei~A
  Efros}.} \bibinfo{year}{2018}\natexlab{}.
\newblock \showarticletitle{Everybody dance now}. In \bibinfo{booktitle}{{\em
  ECCV Workshop}}.
\newblock


\bibitem[\protect\citeauthoryear{Chen, Xu, Yang, and Tao}{Chen
  et~al\mbox{.}}{2018}]%
        {chen2018attention}
\bibfield{author}{\bibinfo{person}{Xinyuan Chen}, \bibinfo{person}{Chang Xu},
  \bibinfo{person}{Xiaokang Yang}, {and} \bibinfo{person}{Dacheng Tao}.}
  \bibinfo{year}{2018}\natexlab{}.
\newblock \showarticletitle{Attention-GAN for object transfiguration in wild
  images}. In \bibinfo{booktitle}{{\em ECCV}}.
\newblock


\bibitem[\protect\citeauthoryear{Choi, Choi, Kim, Ha, Kim, and Choo}{Choi
  et~al\mbox{.}}{2018}]%
        {choi2017stargan}
\bibfield{author}{\bibinfo{person}{Yunjey Choi}, \bibinfo{person}{Minje Choi},
  \bibinfo{person}{Munyoung Kim}, \bibinfo{person}{Jung-Woo Ha},
  \bibinfo{person}{Sunghun Kim}, {and} \bibinfo{person}{Jaegul Choo}.}
  \bibinfo{year}{2018}\natexlab{}.
\newblock \showarticletitle{StarGAN: Unified Generative Adversarial Networks
  for Multi-Domain Image-to-Image Translation}. In \bibinfo{booktitle}{{\em
  CVPR}}.
\newblock


\bibitem[\protect\citeauthoryear{Di, Sindagi, and Patel}{Di
  et~al\mbox{.}}{2018}]%
        {di2017gp}
\bibfield{author}{\bibinfo{person}{Xing Di}, \bibinfo{person}{Vishwanath~A
  Sindagi}, {and} \bibinfo{person}{Vishal~M Patel}.}
  \bibinfo{year}{2018}\natexlab{}.
\newblock \showarticletitle{GP-GAN: gender preserving GAN for synthesizing
  faces from landmarks}. In \bibinfo{booktitle}{{\em ICPR}}.
\newblock


\bibitem[\protect\citeauthoryear{Dolhansky and Canton~Ferrer}{Dolhansky and
  Canton~Ferrer}{2018}]%
        {dolhansky2018eye}
\bibfield{author}{\bibinfo{person}{Brian Dolhansky} {and}
  \bibinfo{person}{Cristian Canton~Ferrer}.} \bibinfo{year}{2018}\natexlab{}.
\newblock \showarticletitle{Eye in-painting with exemplar generative
  adversarial networks}. In \bibinfo{booktitle}{{\em CVPR}}.
\newblock


\bibitem[\protect\citeauthoryear{Duan, Wang, Tang, Latapie, and Yan}{Duan
  et~al\mbox{.}}{2019}]%
        {duan2019cascade}
\bibfield{author}{\bibinfo{person}{Bin Duan}, \bibinfo{person}{Wei Wang},
  \bibinfo{person}{Hao Tang}, \bibinfo{person}{Hugo Latapie}, {and}
  \bibinfo{person}{Yan Yan}.} \bibinfo{year}{2019}\natexlab{}.
\newblock \showarticletitle{Cascade Attention Guided Residue Learning GAN for
  Cross-Modal Translation}.
\newblock \bibinfo{journal}{{\em arXiv preprint arXiv:1907.01826\/}}
  (\bibinfo{year}{2019}).
\newblock


\bibitem[\protect\citeauthoryear{Goodfellow, Pouget-Abadie, Mirza, Xu,
  Warde-Farley, Ozair, Courville, and Bengio}{Goodfellow et~al\mbox{.}}{2014}]%
        {goodfellow2014generative}
\bibfield{author}{\bibinfo{person}{Ian Goodfellow}, \bibinfo{person}{Jean
  Pouget-Abadie}, \bibinfo{person}{Mehdi Mirza}, \bibinfo{person}{Bing Xu},
  \bibinfo{person}{David Warde-Farley}, \bibinfo{person}{Sherjil Ozair},
  \bibinfo{person}{Aaron Courville}, {and} \bibinfo{person}{Yoshua Bengio}.}
  \bibinfo{year}{2014}\natexlab{}.
\newblock \showarticletitle{Generative adversarial nets}. In
  \bibinfo{booktitle}{{\em NIPS}}.
\newblock


\bibitem[\protect\citeauthoryear{Gulrajani, Ahmed, Arjovsky, Dumoulin, and
  Courville}{Gulrajani et~al\mbox{.}}{2017}]%
        {gulrajani2017improved}
\bibfield{author}{\bibinfo{person}{Ishaan Gulrajani}, \bibinfo{person}{Faruk
  Ahmed}, \bibinfo{person}{Martin Arjovsky}, \bibinfo{person}{Vincent
  Dumoulin}, {and} \bibinfo{person}{Aaron Courville}.}
  \bibinfo{year}{2017}\natexlab{}.
\newblock \showarticletitle{Improved training of wasserstein gans}. In
  \bibinfo{booktitle}{{\em NIPS}}.
\newblock


\bibitem[\protect\citeauthoryear{Isola, Zhu, Zhou, and Efros}{Isola
  et~al\mbox{.}}{2017}]%
        {isola2016image}
\bibfield{author}{\bibinfo{person}{Phillip Isola}, \bibinfo{person}{Junyan
  Zhu}, \bibinfo{person}{Tinghui Zhou}, {and} \bibinfo{person}{Alexei~A
  Efros}.} \bibinfo{year}{2017}\natexlab{}.
\newblock \showarticletitle{Image-to-image translation with conditional
  adversarial networks}. In \bibinfo{booktitle}{{\em CVPR}}.
\newblock


\bibitem[\protect\citeauthoryear{Karras, Aila, Laine, and Lehtinen}{Karras
  et~al\mbox{.}}{2018}]%
        {karras2017progressive}
\bibfield{author}{\bibinfo{person}{Tero Karras}, \bibinfo{person}{Timo Aila},
  \bibinfo{person}{Samuli Laine}, {and} \bibinfo{person}{Jaakko Lehtinen}.}
  \bibinfo{year}{2018}\natexlab{}.
\newblock \showarticletitle{Progressive growing of gans for improved quality,
  stability, and variation}. In \bibinfo{booktitle}{{\em ICLR}}.
\newblock


\bibitem[\protect\citeauthoryear{Karras, Laine, and Aila}{Karras
  et~al\mbox{.}}{2019}]%
        {karras2019style}
\bibfield{author}{\bibinfo{person}{Tero Karras}, \bibinfo{person}{Samuli
  Laine}, {and} \bibinfo{person}{Timo Aila}.} \bibinfo{year}{2019}\natexlab{}.
\newblock \showarticletitle{A style-based generator architecture for generative
  adversarial networks}. In \bibinfo{booktitle}{{\em CVPR}}.
\newblock


\bibitem[\protect\citeauthoryear{Kim, Cha, Kim, Lee, and Kim}{Kim
  et~al\mbox{.}}{2017}]%
        {kim2017learning}
\bibfield{author}{\bibinfo{person}{Taeksoo Kim}, \bibinfo{person}{Moonsu Cha},
  \bibinfo{person}{Hyunsoo Kim}, \bibinfo{person}{Jungkwon Lee}, {and}
  \bibinfo{person}{Jiwon Kim}.} \bibinfo{year}{2017}\natexlab{}.
\newblock \showarticletitle{Learning to discover cross-domain relations with
  generative adversarial networks}. In \bibinfo{booktitle}{{\em ICML}}.
\newblock


\bibitem[\protect\citeauthoryear{Kingma and Ba}{Kingma and Ba}{2015}]%
        {kingma2014adam}
\bibfield{author}{\bibinfo{person}{Diederik Kingma} {and}
  \bibinfo{person}{Jimmy Ba}.} \bibinfo{year}{2015}\natexlab{}.
\newblock \showarticletitle{Adam: A method for stochastic optimization}. In
  \bibinfo{booktitle}{{\em ICLR}}.
\newblock


\bibitem[\protect\citeauthoryear{Korshunova, Shi, Dambre, and Theis}{Korshunova
  et~al\mbox{.}}{2017}]%
        {korshunova2016fast}
\bibfield{author}{\bibinfo{person}{Iryna Korshunova}, \bibinfo{person}{Wenzhe
  Shi}, \bibinfo{person}{Joni Dambre}, {and} \bibinfo{person}{Lucas Theis}.}
  \bibinfo{year}{2017}\natexlab{}.
\newblock \showarticletitle{Fast face-swap using convolutional neural
  networks}. In \bibinfo{booktitle}{{\em ICCV}}.
\newblock


\bibitem[\protect\citeauthoryear{Langner, Dotsch, Bijlstra, Wigboldus, Hawk,
  and Van~Knippenberg}{Langner et~al\mbox{.}}{2010}]%
        {langner2010presentation}
\bibfield{author}{\bibinfo{person}{Oliver Langner}, \bibinfo{person}{Ron
  Dotsch}, \bibinfo{person}{Gijsbert Bijlstra}, \bibinfo{person}{Daniel~HJ
  Wigboldus}, \bibinfo{person}{Skyler~T Hawk}, {and} \bibinfo{person}{AD
  Van~Knippenberg}.} \bibinfo{year}{2010}\natexlab{}.
\newblock \showarticletitle{Presentation and validation of the Radboud Faces
  Database}.
\newblock \bibinfo{journal}{{\em Taylor \& Francis Cognition and emotion\/}}
  (\bibinfo{year}{2010}).
\newblock


\bibitem[\protect\citeauthoryear{Ma, Jia, Sun, Schiele, Tuytelaars, and
  Van~Gool}{Ma et~al\mbox{.}}{2017}]%
        {ma2017pose}
\bibfield{author}{\bibinfo{person}{Liqian Ma}, \bibinfo{person}{Xu Jia},
  \bibinfo{person}{Qianru Sun}, \bibinfo{person}{Bernt Schiele},
  \bibinfo{person}{Tinne Tuytelaars}, {and} \bibinfo{person}{Luc Van~Gool}.}
  \bibinfo{year}{2017}\natexlab{}.
\newblock \showarticletitle{Pose guided person image generation}. In
  \bibinfo{booktitle}{{\em NIPS}}.
\newblock


\bibitem[\protect\citeauthoryear{Ma, Sun, Georgoulis, Van~Gool, Schiele, and
  Fritz}{Ma et~al\mbox{.}}{2018}]%
        {ma2018disentangled}
\bibfield{author}{\bibinfo{person}{Liqian Ma}, \bibinfo{person}{Qianru Sun},
  \bibinfo{person}{Stamatios Georgoulis}, \bibinfo{person}{Luc Van~Gool},
  \bibinfo{person}{Bernt Schiele}, {and} \bibinfo{person}{Mario Fritz}.}
  \bibinfo{year}{2018}\natexlab{}.
\newblock \showarticletitle{Disentangled person image generation}. In
  \bibinfo{booktitle}{{\em CVPR}}.
\newblock


\bibitem[\protect\citeauthoryear{Ma, Fu, Wen~Chen, and Mei}{Ma
  et~al\mbox{.}}{2018}]%
        {ma2018gan}
\bibfield{author}{\bibinfo{person}{Shuang Ma}, \bibinfo{person}{Jianlong Fu},
  \bibinfo{person}{Chang Wen~Chen}, {and} \bibinfo{person}{Tao Mei}.}
  \bibinfo{year}{2018}\natexlab{}.
\newblock \showarticletitle{DA-GAN: Instance-level image translation by deep
  attention generative adversarial networks}. In \bibinfo{booktitle}{{\em
  CVPR}}.
\newblock


\bibitem[\protect\citeauthoryear{Mansimov, Parisotto, Ba, and
  Salakhutdinov}{Mansimov et~al\mbox{.}}{2015}]%
        {mansimov2015generating}
\bibfield{author}{\bibinfo{person}{Elman Mansimov}, \bibinfo{person}{Emilio
  Parisotto}, \bibinfo{person}{Jimmy~Lei Ba}, {and} \bibinfo{person}{Ruslan
  Salakhutdinov}.} \bibinfo{year}{2015}\natexlab{}.
\newblock \showarticletitle{Generating images from captions with attention}. In
  \bibinfo{booktitle}{{\em ICLR}}.
\newblock


\bibitem[\protect\citeauthoryear{Mathieu, Couprie, and LeCun}{Mathieu
  et~al\mbox{.}}{2016}]%
        {mathieu2015deep}
\bibfield{author}{\bibinfo{person}{Michael Mathieu}, \bibinfo{person}{Camille
  Couprie}, {and} \bibinfo{person}{Yann LeCun}.}
  \bibinfo{year}{2016}\natexlab{}.
\newblock \showarticletitle{Deep multi-scale video prediction beyond mean
  square error}. In \bibinfo{booktitle}{{\em ICLR}}.
\newblock


\bibitem[\protect\citeauthoryear{Mejjati, Richardt, Tompkin, Cosker, and
  Kim}{Mejjati et~al\mbox{.}}{2018}]%
        {mejjati2018unsupervised}
\bibfield{author}{\bibinfo{person}{Youssef~Alami Mejjati},
  \bibinfo{person}{Christian Richardt}, \bibinfo{person}{James Tompkin},
  \bibinfo{person}{Darren Cosker}, {and} \bibinfo{person}{Kwang~In Kim}.}
  \bibinfo{year}{2018}\natexlab{}.
\newblock \showarticletitle{Unsupervised attention-guided image-to-image
  translation}. In \bibinfo{booktitle}{{\em NeurIPS}}.
\newblock


\bibitem[\protect\citeauthoryear{Mirza and Osindero}{Mirza and
  Osindero}{2014}]%
        {mirza2014conditional}
\bibfield{author}{\bibinfo{person}{Mehdi Mirza} {and} \bibinfo{person}{Simon
  Osindero}.} \bibinfo{year}{2014}\natexlab{}.
\newblock \showarticletitle{Conditional generative adversarial nets}.
\newblock \bibinfo{journal}{{\em arXiv preprint arXiv:1411.1784\/}}
  (\bibinfo{year}{2014}).
\newblock


\bibitem[\protect\citeauthoryear{Mo, Cho, and Shin}{Mo et~al\mbox{.}}{2019}]%
        {mo2018instagan}
\bibfield{author}{\bibinfo{person}{Sangwoo Mo}, \bibinfo{person}{Minsu Cho},
  {and} \bibinfo{person}{Jinwoo Shin}.} \bibinfo{year}{2019}\natexlab{}.
\newblock \showarticletitle{InstaGAN: Instance-aware Image-to-Image
  Translation}. In \bibinfo{booktitle}{{\em ICLR}}.
\newblock


\bibitem[\protect\citeauthoryear{Odena}{Odena}{2016}]%
        {odena2016semi}
\bibfield{author}{\bibinfo{person}{Augustus Odena}.}
  \bibinfo{year}{2016}\natexlab{}.
\newblock \showarticletitle{Semi-supervised learning with generative
  adversarial networks}. In \bibinfo{booktitle}{{\em ICML Workshop}}.
\newblock


\bibitem[\protect\citeauthoryear{Park, Liu, Wang, and Zhu}{Park
  et~al\mbox{.}}{2019}]%
        {park2019semantic}
\bibfield{author}{\bibinfo{person}{Taesung Park}, \bibinfo{person}{Ming-Yu
  Liu}, \bibinfo{person}{Ting-Chun Wang}, {and} \bibinfo{person}{Jun-Yan Zhu}.}
  \bibinfo{year}{2019}\natexlab{}.
\newblock \showarticletitle{Semantic image synthesis with spatially-adaptive
  normalization}. In \bibinfo{booktitle}{{\em CVPR}}.
\newblock


\bibitem[\protect\citeauthoryear{Perarnau, van~de Weijer, Raducanu, and
  {\'A}lvarez}{Perarnau et~al\mbox{.}}{2016}]%
        {perarnau2016invertible}
\bibfield{author}{\bibinfo{person}{Guim Perarnau}, \bibinfo{person}{Joost
  van~de Weijer}, \bibinfo{person}{Bogdan Raducanu}, {and}
  \bibinfo{person}{Jose~M {\'A}lvarez}.} \bibinfo{year}{2016}\natexlab{}.
\newblock \showarticletitle{Invertible Conditional GANs for image editing}. In
  \bibinfo{booktitle}{{\em NIPS Workshop}}.
\newblock


\bibitem[\protect\citeauthoryear{Qiao, Yao, Jiao, Li, Chen, and Wang}{Qiao
  et~al\mbox{.}}{2018}]%
        {qiao2018geometry}
\bibfield{author}{\bibinfo{person}{Fengchun Qiao}, \bibinfo{person}{Naiming
  Yao}, \bibinfo{person}{Zirui Jiao}, \bibinfo{person}{Zhihao Li},
  \bibinfo{person}{Hui Chen}, {and} \bibinfo{person}{Hongan Wang}.}
  \bibinfo{year}{2018}\natexlab{}.
\newblock \showarticletitle{Geometry-Contrastive Generative Adversarial Network
  for Facial Expression Synthesis}.
\newblock \bibinfo{journal}{{\em arXiv preprint arXiv:1802.01822\/}}
  (\bibinfo{year}{2018}).
\newblock


\bibitem[\protect\citeauthoryear{Reed, Akata, Yan, Logeswaran, Schiele, and
  Lee}{Reed et~al\mbox{.}}{2016b}]%
        {reed2016generative}
\bibfield{author}{\bibinfo{person}{Scott Reed}, \bibinfo{person}{Zeynep Akata},
  \bibinfo{person}{Xinchen Yan}, \bibinfo{person}{Lajanugen Logeswaran},
  \bibinfo{person}{Bernt Schiele}, {and} \bibinfo{person}{Honglak Lee}.}
  \bibinfo{year}{2016}\natexlab{b}.
\newblock \showarticletitle{Generative Adversarial Text-to-Image Synthesis}. In
  \bibinfo{booktitle}{{\em ICML}}.
\newblock


\bibitem[\protect\citeauthoryear{Reed, van~den Oord, Kalchbrenner, Bapst,
  Botvinick, and de~Freitas}{Reed et~al\mbox{.}}{2016c}]%
        {reed2016generating}
\bibfield{author}{\bibinfo{person}{Scott Reed}, \bibinfo{person}{A{\"a}ron
  van~den Oord}, \bibinfo{person}{Nal Kalchbrenner}, \bibinfo{person}{Victor
  Bapst}, \bibinfo{person}{Matt Botvinick}, {and} \bibinfo{person}{Nando de
  Freitas}.} \bibinfo{year}{2016}\natexlab{c}.
\newblock \showarticletitle{Generating interpretable images with controllable
  structure}.
\newblock \bibinfo{journal}{{\em Technical Report\/}} (\bibinfo{year}{2016}).
\newblock


\bibitem[\protect\citeauthoryear{Reed, Akata, Mohan, Tenka, Schiele, and
  Lee}{Reed et~al\mbox{.}}{2016a}]%
        {reed2016learning}
\bibfield{author}{\bibinfo{person}{Scott~E Reed}, \bibinfo{person}{Zeynep
  Akata}, \bibinfo{person}{Santosh Mohan}, \bibinfo{person}{Samuel Tenka},
  \bibinfo{person}{Bernt Schiele}, {and} \bibinfo{person}{Honglak Lee}.}
  \bibinfo{year}{2016}\natexlab{a}.
\newblock \showarticletitle{Learning what and where to draw}. In
  \bibinfo{booktitle}{{\em NIPS}}.
\newblock


\bibitem[\protect\citeauthoryear{Regmi and Borji}{Regmi and Borji}{2018}]%
        {regmi2018cross}
\bibfield{author}{\bibinfo{person}{Krishna Regmi} {and} \bibinfo{person}{Ali
  Borji}.} \bibinfo{year}{2018}\natexlab{}.
\newblock \showarticletitle{Cross-view image synthesis using conditional gans}.
  In \bibinfo{booktitle}{{\em CVPR}}.
\newblock


\bibitem[\protect\citeauthoryear{Ronneberger, Fischer, and Brox}{Ronneberger
  et~al\mbox{.}}{2015}]%
        {ronneberger2015u}
\bibfield{author}{\bibinfo{person}{Olaf Ronneberger}, \bibinfo{person}{Philipp
  Fischer}, {and} \bibinfo{person}{Thomas Brox}.}
  \bibinfo{year}{2015}\natexlab{}.
\newblock \showarticletitle{U-net: Convolutional networks for biomedical image
  segmentation}. In \bibinfo{booktitle}{{\em MICCAI}}.
\newblock


\bibitem[\protect\citeauthoryear{Salimans, Goodfellow, Zaremba, Cheung,
  Radford, and Chen}{Salimans et~al\mbox{.}}{2016}]%
        {salimans2016improved}
\bibfield{author}{\bibinfo{person}{Tim Salimans}, \bibinfo{person}{Ian
  Goodfellow}, \bibinfo{person}{Wojciech Zaremba}, \bibinfo{person}{Vicki
  Cheung}, \bibinfo{person}{Alec Radford}, {and} \bibinfo{person}{Xi Chen}.}
  \bibinfo{year}{2016}\natexlab{}.
\newblock \showarticletitle{Improved techniques for training gans}. In
  \bibinfo{booktitle}{{\em NIPS}}.
\newblock


\bibitem[\protect\citeauthoryear{Siarohin, Lathuili{\`e}re, Tulyakov, Ricci,
  and Sebe}{Siarohin et~al\mbox{.}}{2019}]%
        {siarohin2019animating}
\bibfield{author}{\bibinfo{person}{Aliaksandr Siarohin},
  \bibinfo{person}{St{\'e}phane Lathuili{\`e}re}, \bibinfo{person}{Sergey
  Tulyakov}, \bibinfo{person}{Elisa Ricci}, {and} \bibinfo{person}{Nicu Sebe}.}
  \bibinfo{year}{2019}\natexlab{}.
\newblock \showarticletitle{Animating arbitrary objects via deep motion
  transfer}. In \bibinfo{booktitle}{{\em CVPR}}.
\newblock


\bibitem[\protect\citeauthoryear{Siarohin, Sangineto, Lathuiliere, and
  Sebe}{Siarohin et~al\mbox{.}}{2018}]%
        {siarohin2017deformable}
\bibfield{author}{\bibinfo{person}{Aliaksandr Siarohin}, \bibinfo{person}{Enver
  Sangineto}, \bibinfo{person}{Stephane Lathuiliere}, {and}
  \bibinfo{person}{Nicu Sebe}.} \bibinfo{year}{2018}\natexlab{}.
\newblock \showarticletitle{Deformable GANs for Pose-based Human Image
  Generation}. In \bibinfo{booktitle}{{\em CVPR}}.
\newblock


\bibitem[\protect\citeauthoryear{Song, Lu, He, Sun, and Tan}{Song
  et~al\mbox{.}}{2018}]%
        {song2017geometry}
\bibfield{author}{\bibinfo{person}{Lingxiao Song}, \bibinfo{person}{Zhihe Lu},
  \bibinfo{person}{Ran He}, \bibinfo{person}{Zhenan Sun}, {and}
  \bibinfo{person}{Tieniu Tan}.} \bibinfo{year}{2018}\natexlab{}.
\newblock \showarticletitle{Geometry guided adversarial facial expression
  synthesis}. In \bibinfo{booktitle}{{\em ACM MM}}.
\newblock


\bibitem[\protect\citeauthoryear{Sun, Ma, Joon~Oh, Van~Gool, Schiele, and
  Fritz}{Sun et~al\mbox{.}}{2018}]%
        {sun2018natural}
\bibfield{author}{\bibinfo{person}{Qianru Sun}, \bibinfo{person}{Liqian Ma},
  \bibinfo{person}{Seong Joon~Oh}, \bibinfo{person}{Luc Van~Gool},
  \bibinfo{person}{Bernt Schiele}, {and} \bibinfo{person}{Mario Fritz}.}
  \bibinfo{year}{2018}\natexlab{}.
\newblock \showarticletitle{Natural and effective obfuscation by head
  inpainting}. In \bibinfo{booktitle}{{\em CVPR}}.
\newblock


\bibitem[\protect\citeauthoryear{Taigman, Polyak, and Wolf}{Taigman
  et~al\mbox{.}}{2017}]%
        {taigman2016unsupervised}
\bibfield{author}{\bibinfo{person}{Yaniv Taigman}, \bibinfo{person}{Adam
  Polyak}, {and} \bibinfo{person}{Lior Wolf}.} \bibinfo{year}{2017}\natexlab{}.
\newblock \showarticletitle{Unsupervised cross-domain image generation}. In
  \bibinfo{booktitle}{{\em ICLR}}.
\newblock


\bibitem[\protect\citeauthoryear{Tang, Wang, Xu, Yan, and Sebe}{Tang
  et~al\mbox{.}}{2018}]%
        {tang2018gesturegan}
\bibfield{author}{\bibinfo{person}{Hao Tang}, \bibinfo{person}{Wei Wang},
  \bibinfo{person}{Dan Xu}, \bibinfo{person}{Yan Yan}, {and}
  \bibinfo{person}{Nicu Sebe}.} \bibinfo{year}{2018}\natexlab{}.
\newblock \showarticletitle{GestureGAN for Hand Gesture-to-Gesture Translation
  in the Wild}. In \bibinfo{booktitle}{{\em ACM MM}}.
\newblock


\bibitem[\protect\citeauthoryear{Tang, Xu, Sebe, Wang, Corso, and Yan}{Tang
  et~al\mbox{.}}{2019b}]%
        {tang2019multi}
\bibfield{author}{\bibinfo{person}{Hao Tang}, \bibinfo{person}{Dan Xu},
  \bibinfo{person}{Nicu Sebe}, \bibinfo{person}{Yanzhi Wang},
  \bibinfo{person}{Jason~J Corso}, {and} \bibinfo{person}{Yan Yan}.}
  \bibinfo{year}{2019}\natexlab{b}.
\newblock \showarticletitle{Multi-channel attention selection gan with cascaded
  semantic guidance for cross-view image translation}. In
  \bibinfo{booktitle}{{\em CVPR}}.
\newblock


\bibitem[\protect\citeauthoryear{Tang, Xu, Sebe, and Yan}{Tang
  et~al\mbox{.}}{2019a}]%
        {tang2019attention}
\bibfield{author}{\bibinfo{person}{Hao Tang}, \bibinfo{person}{Dan Xu},
  \bibinfo{person}{Nicu Sebe}, {and} \bibinfo{person}{Yan Yan}.}
  \bibinfo{year}{2019}\natexlab{a}.
\newblock \showarticletitle{Attention-Guided Generative Adversarial Networks
  for Unsupervised Image-to-Image Translation}. In \bibinfo{booktitle}{{\em
  IJCNN}}.
\newblock


\bibitem[\protect\citeauthoryear{Tang, Xu, Wang, Yan, and Sebe}{Tang
  et~al\mbox{.}}{2018}]%
        {tang2018dual}
\bibfield{author}{\bibinfo{person}{Hao Tang}, \bibinfo{person}{Dan Xu},
  \bibinfo{person}{Wei Wang}, \bibinfo{person}{Yan Yan}, {and}
  \bibinfo{person}{Nicu Sebe}.} \bibinfo{year}{2018}\natexlab{}.
\newblock \showarticletitle{Dual Generator Generative Adversarial Networks for
  Multi-Domain Image-to-Image Translation}. In \bibinfo{booktitle}{{\em ACCV}}.
\newblock


\bibitem[\protect\citeauthoryear{Van Den~Oord, Dieleman, Zen, Simonyan,
  Vinyals, Graves, Kalchbrenner, Senior, and Kavukcuoglu}{Van Den~Oord
  et~al\mbox{.}}{2016}]%
        {oord2016wavenet}
\bibfield{author}{\bibinfo{person}{A{\"a}ron Van Den~Oord},
  \bibinfo{person}{Sander Dieleman}, \bibinfo{person}{Heiga Zen},
  \bibinfo{person}{Karen Simonyan}, \bibinfo{person}{Oriol Vinyals},
  \bibinfo{person}{Alex Graves}, \bibinfo{person}{Nal Kalchbrenner},
  \bibinfo{person}{Andrew~W Senior}, {and} \bibinfo{person}{Koray
  Kavukcuoglu}.} \bibinfo{year}{2016}\natexlab{}.
\newblock \showarticletitle{WaveNet: A generative model for raw audio.}. In
  \bibinfo{booktitle}{{\em SSW}}.
\newblock


\bibitem[\protect\citeauthoryear{Vondrick, Pirsiavash, and Torralba}{Vondrick
  et~al\mbox{.}}{2016}]%
        {vondrick2016generating}
\bibfield{author}{\bibinfo{person}{Carl Vondrick}, \bibinfo{person}{Hamed
  Pirsiavash}, {and} \bibinfo{person}{Antonio Torralba}.}
  \bibinfo{year}{2016}\natexlab{}.
\newblock \showarticletitle{Generating videos with scene dynamics}. In
  \bibinfo{booktitle}{{\em NIPS}}.
\newblock


\bibitem[\protect\citeauthoryear{Wang, Liu, Zhu, Liu, Tao, Kautz, and
  Catanzaro}{Wang et~al\mbox{.}}{2018a}]%
        {wang2018video}
\bibfield{author}{\bibinfo{person}{Tingchun Wang}, \bibinfo{person}{Mingyu
  Liu}, \bibinfo{person}{Junyan Zhu}, \bibinfo{person}{Guilin Liu},
  \bibinfo{person}{Andrew Tao}, \bibinfo{person}{Jan Kautz}, {and}
  \bibinfo{person}{Bryan Catanzaro}.} \bibinfo{year}{2018}\natexlab{a}.
\newblock \showarticletitle{Video-to-video synthesis}. In
  \bibinfo{booktitle}{{\em NeurIPS}}.
\newblock


\bibitem[\protect\citeauthoryear{Wang, Liu, Zhu, Tao, Kautz, and
  Catanzaro}{Wang et~al\mbox{.}}{2018b}]%
        {wang2018pix2pixHD}
\bibfield{author}{\bibinfo{person}{Tingchun Wang}, \bibinfo{person}{Mingyu
  Liu}, \bibinfo{person}{Junyan Zhu}, \bibinfo{person}{Andrew Tao},
  \bibinfo{person}{Jan Kautz}, {and} \bibinfo{person}{Bryan Catanzaro}.}
  \bibinfo{year}{2018}\natexlab{b}.
\newblock \showarticletitle{High-Resolution Image Synthesis and Semantic
  Manipulation with Conditional GANs}. In \bibinfo{booktitle}{{\em CVPR}}.
\newblock


\bibitem[\protect\citeauthoryear{Wang, Alameda-Pineda, Xu, Fua, Ricci, and
  Sebe}{Wang et~al\mbox{.}}{2018}]%
        {wang2018every}
\bibfield{author}{\bibinfo{person}{Wei Wang}, \bibinfo{person}{Xavier
  Alameda-Pineda}, \bibinfo{person}{Dan Xu}, \bibinfo{person}{Pascal Fua},
  \bibinfo{person}{Elisa Ricci}, {and} \bibinfo{person}{Nicu Sebe}.}
  \bibinfo{year}{2018}\natexlab{}.
\newblock \showarticletitle{Every smile is unique: Landmark-guided diverse
  smile generation}. In \bibinfo{booktitle}{{\em CVPR}}.
\newblock


\bibitem[\protect\citeauthoryear{Wang and Gupta}{Wang and Gupta}{2016}]%
        {wang2016generative}
\bibfield{author}{\bibinfo{person}{Xiaolong Wang} {and}
  \bibinfo{person}{Abhinav Gupta}.} \bibinfo{year}{2016}\natexlab{}.
\newblock \showarticletitle{Generative image modeling using style and structure
  adversarial networks}. In \bibinfo{booktitle}{{\em ECCV}}.
\newblock


\bibitem[\protect\citeauthoryear{Wang, Bovik, Sheikh, and Simoncelli}{Wang
  et~al\mbox{.}}{2004}]%
        {wang2004image}
\bibfield{author}{\bibinfo{person}{Zhou Wang}, \bibinfo{person}{Alan~C Bovik},
  \bibinfo{person}{Hamid~R Sheikh}, {and} \bibinfo{person}{Eero~P Simoncelli}.}
  \bibinfo{year}{2004}\natexlab{}.
\newblock \showarticletitle{Image quality assessment: from error visibility to
  structural similarity}.
\newblock \bibinfo{journal}{{\em IEEE TIP\/}} \bibinfo{volume}{13},
  \bibinfo{number}{4} (\bibinfo{year}{2004}), \bibinfo{pages}{600--612}.
\newblock


\bibitem[\protect\citeauthoryear{Yan, Xu, Ni, Zhang, and Yang}{Yan
  et~al\mbox{.}}{2017}]%
        {yan2017skeleton}
\bibfield{author}{\bibinfo{person}{Yichao Yan}, \bibinfo{person}{Jingwei Xu},
  \bibinfo{person}{Bingbing Ni}, \bibinfo{person}{Wendong Zhang}, {and}
  \bibinfo{person}{Xiaokang Yang}.} \bibinfo{year}{2017}\natexlab{}.
\newblock \showarticletitle{Skeleton-aided Articulated Motion Generation}. In
  \bibinfo{booktitle}{{\em ACM MM}}.
\newblock


\bibitem[\protect\citeauthoryear{Yi, Zhang, Gong, et~al\mbox{.}}{Yi
  et~al\mbox{.}}{2017}]%
        {yi2017dualgan}
\bibfield{author}{\bibinfo{person}{Zili Yi}, \bibinfo{person}{Hao Zhang},
  \bibinfo{person}{Ping~Tan Gong}, {et~al\mbox{.}}}
  \bibinfo{year}{2017}\natexlab{}.
\newblock \showarticletitle{DualGAN: Unsupervised Dual Learning for
  Image-to-Image Translation}. In \bibinfo{booktitle}{{\em ICCV}}.
\newblock


\bibitem[\protect\citeauthoryear{Yu, Zhang, Wang, and Yu}{Yu
  et~al\mbox{.}}{2017}]%
        {yu2017seqgan}
\bibfield{author}{\bibinfo{person}{Lantao Yu}, \bibinfo{person}{Weinan Zhang},
  \bibinfo{person}{Jun Wang}, {and} \bibinfo{person}{Yong Yu}.}
  \bibinfo{year}{2017}\natexlab{}.
\newblock \showarticletitle{SeqGAN: Sequence Generative Adversarial Nets with
  Policy Gradient.}. In \bibinfo{booktitle}{{\em AAAI}}.
\newblock


\bibitem[\protect\citeauthoryear{Zhang, Sun, Chen, Tang, Yan, Qin, and
  Sebe}{Zhang et~al\mbox{.}}{2019}]%
        {zhang2019gazecorrection}
\bibfield{author}{\bibinfo{person}{Jichao Zhang}, \bibinfo{person}{Meng Sun},
  \bibinfo{person}{Jingjing Chen}, \bibinfo{person}{Hao Tang},
  \bibinfo{person}{Yan Yan}, \bibinfo{person}{Xueying Qin}, {and}
  \bibinfo{person}{Nicu Sebe}.} \bibinfo{year}{2019}\natexlab{}.
\newblock \showarticletitle{GazeCorrection: Self-Guided Eye Manipulation in the
  wild using Self-Supervised Generative Adversarial Networks}.
\newblock \bibinfo{journal}{{\em arXiv preprint arXiv:1906.00805\/}}
  (\bibinfo{year}{2019}).
\newblock


\bibitem[\protect\citeauthoryear{Zheng, Shen, Tian, Wang, Wang, and Tian}{Zheng
  et~al\mbox{.}}{2015}]%
        {zheng2015scalable}
\bibfield{author}{\bibinfo{person}{Liang Zheng}, \bibinfo{person}{Liyue Shen},
  \bibinfo{person}{Lu Tian}, \bibinfo{person}{Shengjin Wang},
  \bibinfo{person}{Jingdong Wang}, {and} \bibinfo{person}{Qi Tian}.}
  \bibinfo{year}{2015}\natexlab{}.
\newblock \showarticletitle{Scalable person re-identification: A benchmark}. In
  \bibinfo{booktitle}{{\em ICCV}}.
\newblock


\bibitem[\protect\citeauthoryear{Zhou, Xiao, Yang, Feng, He, and He}{Zhou
  et~al\mbox{.}}{2017}]%
        {zhou2017genegan}
\bibfield{author}{\bibinfo{person}{Shuchang Zhou}, \bibinfo{person}{Taihong
  Xiao}, \bibinfo{person}{Yi Yang}, \bibinfo{person}{Dieqiao Feng},
  \bibinfo{person}{Qinyao He}, {and} \bibinfo{person}{Weiran He}.}
  \bibinfo{year}{2017}\natexlab{}.
\newblock \showarticletitle{GeneGAN: Learning Object Transfiguration and
  Attribute Subspace from Unpaired Data}. In \bibinfo{booktitle}{{\em BMVC}}.
\newblock


\bibitem[\protect\citeauthoryear{Zhu, Park, Isola, and Efros}{Zhu
  et~al\mbox{.}}{2017}]%
        {zhu2017unpaired}
\bibfield{author}{\bibinfo{person}{Junyan Zhu}, \bibinfo{person}{Taesung Park},
  \bibinfo{person}{Phillip Isola}, {and} \bibinfo{person}{Alexei~A Efros}.}
  \bibinfo{year}{2017}\natexlab{}.
\newblock \showarticletitle{Unpaired image-to-image translation using
  cycle-consistent adversarial networks}. In \bibinfo{booktitle}{{\em ICCV}}.
\newblock


\end{thebibliography}

\end{document}